\pdfoutput=1

\documentclass[10pt,twocolumn,letterpaper]{article}

\usepackage[pagenumbers]{cvpr}      %

\PassOptionsToPackage{noabbrev}{cleveref}
\usepackage{xspace}
\usepackage{times,latexsym}
\usepackage[T1]{fontenc}
\usepackage[utf8]{inputenc}
\usepackage{microtype,inconsolata}
\usepackage{fvextra}
\usepackage{graphicx,booktabs,tabularx}
\usepackage{subcaption}
\usepackage{amsmath,amssymb}
\usepackage{xcolor,tikz}
\usepackage{enumitem}

\usetikzlibrary{arrows.meta,positioning,fit}
\definecolor{pendinggray}{gray}{0.55}
\definecolor{methodblue}{RGB}{35,83,112}

\newcommand{\abstain}{\texttt{cannot be determined}}
\newcommand{\ind}{\mathbf{1}}

\newif\ifshowcomments

\showcommentsfalse

\ifshowcomments
  \newcommand{\zy}[1]{\textcolor{brown}{[zhaoyu: #1]}}
  \newcommand{\rs}[1]{\textcolor{blue}{[rs: #1]}}
\else
  \newcommand{\zy}[1]{\unskip}
  \newcommand{\rs}[1]{\unskip}
\fi

\newcommand{\MethodName}{\textsc{FaithTwI}\xspace}
\AtEndPreamble{%
  \crefname{section}{Section}{Sections}%
  \crefname{table}{Table}{Tables}%
}

\definecolor{cvprblue}{rgb}{0.21,0.49,0.74}
\usepackage[pagebackref,breaklinks,colorlinks,allcolors=cvprblue]{hyperref}

\def\confName{CVPR}
\def\confYear{2027}

\title{Learning to Think with Images Faithfully via Adaptive Data Selection}

\author{
Yu Zhao$^{1,*}$ \quad Jiaxin Zhang$^{2,\dagger}$ \quad Rohit Saxena$^{1}$ \quad
Fazl Barez$^{3}$ \quad Pasquale Minervini$^{1,5}$ \quad Yiwen Guo$^{4,\dagger}$ \\[0.6em]
$^{1}$University of Edinburgh \quad $^{2}$LIGHTSPEED \quad $^{3}$University of Oxford \\
$^{4}$Independent Researcher \quad $^{5}$Miniml.AI \\[0.4em]
{\small $^{*}$\texttt{yu.zhao@ed.ac.uk} \quad $^{\dagger}$Corresponding author}
}

\begin{document}
\maketitle
\begin{abstract}
Thinking with images (TwI) empowers vision--language models (VLMs) to overcome perceptual bottlenecks by actively cropping and zooming into fine-grained visual details, promising both higher accuracy and interpretable, region-grounded reasoning.
However, our analysis shows that the improved accuracy of TwI models does not guarantee faithful tool use; they frequently invoke tools without actually relying on the zoomed region, undermining the reliability of tool-assisted reasoning.
We trace this unfaithful behaviour to the training data: when questions are answerable without zooming, the model can answer correctly while ignoring the returned evidence, and this problem is exacerbated during training as the model becomes better at answering without tools.
To keep zooming useful throughout training, we adapt the training data to the model's evolving capability and train on questions where correct zoomed evidence can still help the model answer correctly.
This selection is supported by a data synthesis pipeline that localises small objects and constructs verifiable questions about their fine details, and complemented by an adaptiveness reward that favours tool-free answers when zooming is unnecessary.
The experimental results show that our method substantially improves reasoning faithfulness and achieves state-of-the-art performance on visual search benchmarks ($+11.6\%$ on V*, $+7.9\%$ on HR-Bench 4K, and $+9.4\%$ on HR-Bench 8K).
On multimodal hallucination benchmarks, our model also outperforms the compared TwI baselines.
These results suggest that improving TwI faithfulness does not sacrifice accuracy and that both can be learnt from the right training data without directly supervising the reasoning process.
\end{abstract}

\section{Introduction}

\begin{figure}[!t]
\centering
\includegraphics[width=\linewidth]{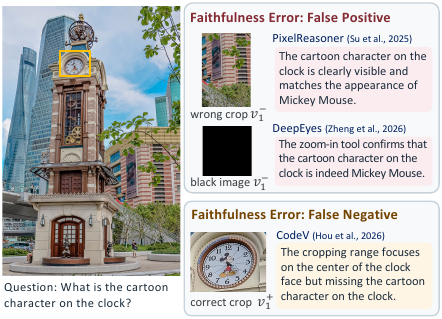}
\caption{Two faithfulness failure modes in existing TwI models.
\textit{False positive}: replacing the tool-returned crop with a random and a black image, PixelReasoner~\citep{wang2025pixelreasoner} and DeepEyes~\citep{zheng2025deepeyes} claim to see the target in the crop.
\textit{False negative}: CodeV~\citep{you2025codev} predicts a correct area but denies the target in it, though its answer is correct after additional zooms.
All three models answer Mickey Mouse correctly but are not faithfully grounded on the evidence.}
\label{fig:teaser}
\end{figure}

Vision-language models (VLMs) often struggle to recognise small objects, dense text, and fine-grained spatial relationships from an image~\citep{lai2025minio3,bai2025qwen25vl,zheng2025deepeyes}.
Thinking with images \citep[TwI;][]{openai2025o3} mitigates this perceptual bottleneck by enabling VLMs to act as multimodal agents that actively crop and zoom into regions of interest and perform interleaved reasoning on the text and zoomed images~\citep{openai2025o3,zheng2025deepeyes,wang2025pixelreasoner,you2025codev,lai2025minio3,zhang2025adaptivecof,hong2025deepeyesv2}.
Beyond improving answer accuracy, TwI facilitates explainable, region-grounded question answering, where each zoomed image provides explicit visual evidence intended to support the model's prediction.
However, this explainability holds only when the model faithfully interprets the images returned by its tools.~\citep{ma2026med,guo2026benefit,shao2026textcall,wang2026illusion,zhang2026insensitivity,li2026viebench}.

In this work, we first examine the faithfulness of TwI models: a model is faithful if it generates reasoning traces and outputs consistent with the content of the zoomed images.
For example, a faithful model should accurately recognise the objects in the zoomed images without hallucinating objects that are not present.
Faithfulness is thus distinct from answer accuracy: a model that zooms into the wrong region can still behave faithfully, while a correct answer can be based on an unsupported visual claim.
We audit the outputs of DeepEyes~\citep{zheng2025deepeyes}, PixelReasoner~\citep{wang2025pixelreasoner}, and CodeV~\citep{you2025codev} and identify two distinct faithfulness failure modes (\cref{fig:teaser}).
Specifically, DeepEyes and PixelReasoner frequently claim to see objects that are absent from a zoomed image in $98.8\%$ of instances.
Conversely, CodeV does not recognise objects that are visible in the zoomed region in $87.9\%$ of instances, even though CodeV aims to improve faithfulness by training the model to zoom into the correct region~\citep{you2025codev}.

To investigate how unfaithful TwI may emerge, we compare model performance with and without tool access and track how this relationship evolves during reinforcement learning (RL) (\cref{sec:diagnosis}).
Although DeepEyes, PixelReasoner, and CodeV call tools on over $94\%$ of questions, zooming improves their accuracy by at most $3.0$ points and sometimes lowers it (\cref{fig:observation-tool-accuracy}).
Moreover, the accuracy of DeepEyes without tools on V*~\citep{gao2024vstar} rises from $71.9\%$ to $83.3\%$ during RL training (\cref{fig:observation-dynamics}).
These observations suggest a potential source of unfaithful tool use: when training questions can be answered correctly without zooming, outcome-based rewards may reinforce correct answers without requiring the model to interpret the returned visual evidence.
This means that during training, the model can learn to answer questions without relying on the tool, making tool use decorative for those instances.
This leads to our hypothesis that unfaithful tool use may stem from training questions that do not require zooming.

Building on this hypothesis, we propose adapting the training data to the model's evolving capabilities.
Specifically, we select questions where providing the correct zoomed evidence improves the current model's answer accuracy compared with using the whole image alone (\cref{sec:selection}).
To support this selection, we build a data synthesis pipeline that uses SAM3~\citep{carion2025sam3} to localise small objects and a VLM to generate questions about their fine details.
The localised regions also provide the zoomed evidence needed for the data selection method (\cref{sec:data-synthesis}).
We further introduce an adaptiveness reward that encourages correct tool-free answers when the model can answer reliably without zooming (\cref{sec:reward}).
This helps the model use its tool-free answering capability, allowing data selection to better identify questions that benefit from zoomed evidence.
Our experiments show that adaptive data selection and the adaptiveness reward jointly improve answer accuracy and faithfulness (\cref{sec:ablations}).

We evaluate our model \MethodName on perception-intensive benchmarks that require inspecting fine-grained visual details in high-resolution images.
The experimental results show that based on Qwen2.5-VL-7B~\citep{bai2025qwen25vl}, our method \MethodName significantly improves the accuracy ($78.5\% \rightarrow 90.1\%$ on V*~\citep{gao2024vstar}, $67.9\% \rightarrow 77.3\%$ on HR-Bench 8K~\citep{ding2024hrbench}, and $42.6\% \rightarrow 56.8\%$ on ZoomBench~\citep{wei2026zwz}) and achieves state-of-the-art performance.
The audit of the reasoning trace shows that \MethodName more accurately recognises whether the target is present in the zoomed region, achieving $94.5\%$ accuracy and surpassing PixelReasoner ($70.8\%$) and CodeV ($55.2\%$). 
On the multimodal hallucination benchmarks HaloQuest~\citep{DBLP:conf/eccv/WangBYLLG24} and VizWiz~\citep{DBLP:conf/cvpr/Gurari0SGLGLB18}, \MethodName achieves the highest scores among the compared TwI models, reaching $70.7\%$ and $57.1\%$ compared with $62.2\%$ and $54.9\%$ for CodeV.
These results further suggest that \MethodName uses the visual evidence more faithfully.

In summary, our contributions are threefold:
\textit{1)} We analyse how TwI models interpret returned visual evidence, tracing unfaithful tool use to training data that can be answered without zooming (\cref{sec:background}).
\textit{2)} We propose a data-driven method to improve accuracy and faithfulness; it adapts the training data to the model's evolving capability, supported by a question synthesis pipeline and an adaptiveness reward (\cref{sec:method}).
\textit{3)} Our experimental results show improvements in both faithfulness and accuracy, with the best accuracy among compared models on visual search and multimodal hallucination benchmarks. (\cref{sec:experiments}).

\section{Evidence of Unfaithful TwI} \label{sec:background}

\paragraph{Background.}
Given an image $I$ and question $q$, a policy $\pi_\theta$ interleaves reasoning text, tool calls, and zoomed images before producing an answer $\hat{y}$.
For a crop-and-zoom tool, the interaction trajectory $\tau$ is:
\begin{equation} \label{eq:trajectory}
\tau = \left( z_1, v_1,\ldots,z_K, v_K,z_{K+1},\hat y \right),
\end{equation}
\noindent where $z_i$ is the $i$-th turn of reasoning, and $v_i$ is the cropped and zoomed region called by $z_i$.
At each step, $\pi_\theta$ generates reasoning and tool requests conditioned on the question $q$, the image $I$, and the interaction history.
The final reasoning turn $z_{K+1}$ outputs $\hat{y}$ without invoking the tool; direct answering, i.e., tool-free answering, corresponds to $K=0$, where $\tau=(z_1,\hat y)$.
For a target specified in $z_i$, the zoomed region $v_i$ either contains the target ($v_i^+$) or misses it ($v_i^-$).
A faithful model should produce subsequent reasoning and action $z_{i+1}$ consistent with the visual evidence returned in $v_i$, e.g., correctly describing $v_i$ and answering the question conditioned on it.

We first audit whether the three TwI models, namely DeepEyes~\citep{zheng2025deepeyes}, PixelReasoner~\citep{wang2025pixelreasoner}, and CodeV~\citep{you2025codev}, faithfully distinguish target-containing crops ($v_1^+$) from crops that miss the target ($v_1^-$).
We then trace how this behaviour and the benefit of tool use evolve during DeepEyes training.

\subsection{Recognising Target Presence and Absence} \label{sec:diagnosis}
We test whether a model's response is consistent with the visual evidence returned by its first tool call on V*~\citep{gao2024vstar}.
To control whether the returned image contains the target, we replace $v_1$ with either a crop centred on the annotated target ($v_1^+$) or a random crop without it ($v_1^-$); we refer to this as the controlled setting in~\cref{tab:evidence-use}.
Then, in the next reasoning turn $z_2$, we check whether the model claims to see the target when it is present and acknowledges its absence when it is not.
We report the \emph{false negative rate} (FNR), which represents how often the model claims not to see the target given $v_1^+$, and the \emph{false positive rate} (FPR), which represents how often it claims to see the target given $v_1^-$.
Recognition accuracy (Acc.) is the percentage of crops for which the response correctly identifies whether the target is present.

\begin{table}[t]
\centering\small
\setlength{\tabcolsep}{2pt}
\resizebox{0.81\linewidth}{!}{%
\begin{tabular}{lrrr@{\hspace{10pt}}rrr}
\toprule
 & \multicolumn{3}{c}{Controlled setting} & \multicolumn{3}{c}{Real-crop setting} \\
\cmidrule(lr){2-4}\cmidrule(lr){5-7}
Model & FNR$\downarrow$ & FPR$\downarrow$ & Acc.$\uparrow$ & FNR$\downarrow$ & FPR$\downarrow$ & Acc.$\uparrow$ \\
\midrule
DeepEyes & $1.6$ & $98.8$ & $50.0$ & $0.0$ & $94.1$ & $73.9$ \\
PixelReasoner & $2.3$ & $56.9$ & $70.8$ & $2.8$ & $91.9$ & $54.6$ \\
CodeV & $87.9$ & $0.7$ & $55.2$ & $84.1$ & $10.7$ & $31.6$ \\
\MethodName & $8.7$ & $2.2$ & $94.5$ & $6.1$ & $18.2$ & $93.3$ \\
\bottomrule
\end{tabular}
}
\caption{Target recognition on V* (\%). The controlled setting (left) replaces the models' first crops with ground-truth crops $v^+$ or target-absent crops $v^-$; the real-crop setting (right) uses the models' own predicted crops, and its details are provided in \cref{app:section3-protocol}.}
\label{tab:evidence-use}
\end{table}

\paragraph{Opposite response biases in TwI models.}
In the controlled setting (\cref{tab:evidence-use}), DeepEyes and PixelReasoner obtain low FNRs ($1.6\%$ and $2.3\%$), but frequently claim to see the target in crops that do not contain it, with FPRs of $98.8\%$ and $56.9\%$.
CodeV shows the opposite bias: its FPR is only $0.7\%$, but it claims not to see the target in $87.9\%$ of the crops that contain it.
Therefore, a low FPR or a low FNR alone does not establish faithful recognition.
Overall, DeepEyes and CodeV reach recognition accuracies of only $50.0\%$ and $55.2\%$, while our model \MethodName reaches $94.5\%$, with both error rates low.
We also evaluate target recognition using the models' own first crops without replacement (real crop in \cref{tab:evidence-use}); more details and analysis are provided in~\cref{app:section3-protocol}.
The above results show whether the first response after zooming is consistent with the returned evidence, and we further analyse how the models behave when the crop misses the target in~\cref{app:human-eval}.

\subsection{Tool-Free Accuracy Improves During Training}
\label{sec:direct-dynamics}
\label{sec:faithfulness-dynamics}
\begin{figure}[t]
\centering
\includegraphics[width=\linewidth]{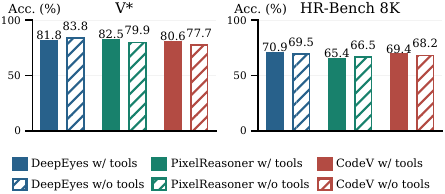}
\caption{Accuracy with and without tools.}
\label{fig:observation-tool-accuracy}

\nextfloat
\vspace{0.8em}
\centering
\begin{subfigure}{0.49\columnwidth}
\centering\includegraphics[width=\linewidth]{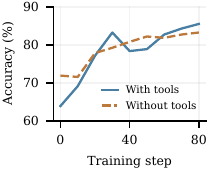}
\caption{Answer accuracy on V*}
\end{subfigure}\hfill
\begin{subfigure}{0.49\columnwidth}
\centering\includegraphics[width=\linewidth]{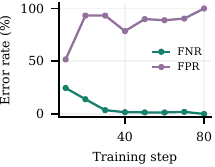}
\caption{FNR and FPR}
\end{subfigure}
\caption{Accuracy and error rate change during DeepEyes training. (a) Answer accuracy improves with and without tools. (b) Responses increasingly claim target presence regardless of the returned crop: FNR falls towards zero while FPR rises to $100\%$.}
\label{fig:observation-dynamics}
\end{figure}

\paragraph{Frequent tool invocation yields minimal accuracy gains.}
Comparing evaluation accuracy with and without tools (\cref{fig:observation-tool-accuracy}) reveals a mismatch between tool usage and utility.
Although DeepEyes, PixelReasoner, and CodeV invoke tools on nearly all queries ($>94\%$), zooming yields modest or even negative returns: DeepEyes drops in accuracy with tools on V* and gains only $+1.4\%$ on HR-Bench 8K; PixelReasoner gains $+2.6\%$ on V* but loses $-1.1\%$ on HR-Bench 8K; and CodeV gains only $+3.0\%$ and $+1.1\%$, respectively.
Recent works also find limited or inconsistent accuracy gains from tool access in TwI~\citep{wang2026illusion,guo2026benefit}.

\begin{figure*}[t]
\centering
\includegraphics[width=\textwidth]{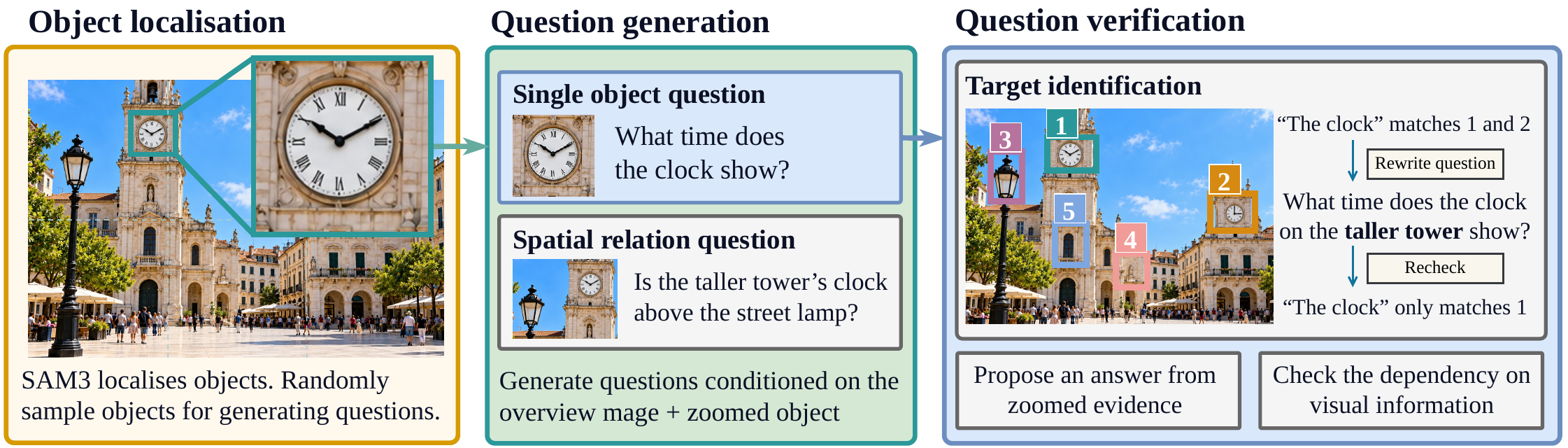}
\caption{Question synthesis workflow. We first localise and sample objects from high-resolution, information-dense images using SAM3~\citep{carion2025sam3}. Then, we generate questions for randomly sampled objects using the image overview and zoomed evidence. Finally, we verify questions in several steps: target identification with Set-of-Mark prompting~\citep{yang2023setofmark} checks if the question refers to unambiguous objects; answer correctness checks if the proposed answer is reliable; visual dependency checks if the question can be guessed without seeing the image.}
\label{fig:method-overview}
\end{figure*}

\paragraph{How decorative tool use emerges during RL training.}
To trace how this mismatch develops, we replicate DeepEyes training and track its performance across checkpoints (\cref{fig:observation-dynamics}).
Remarkably, the model's tool-free accuracy rises substantially during tool-augmented training, from $71.9\%$ to $83.3\%$ on V* and from $69.3\%$ to $73.5\%$ on HR-Bench 4K, while the model invokes tool frequently.
However, the model increasingly claims to see the target regardless of whether it is present: FNR decreases from $24.6\%$ to $0\%$, while FPR increases from $51.7\%$ to $100\%$
By the end of training, the model claims to see the target in every first crop, even when the crop misses it completely.

\paragraph{Hypothesis and implications for training.}
These observations show that higher accuracy does not imply more faithful use of the returned evidence: during training, accuracy without tools rises while the model's claims about the returned crops become less faithful, and the trained model gains little from tool use.
These results suggest that many questions can be answered correctly without zooming, and the model needs zooming for fewer questions as training proceeds.
The model can correctly answer such questions even if it ignores the zoomed crop, so training does not require the model to use the crop faithfully.
This leads us to hypothesise that unfaithful tool use may stem from training questions that do not require zooming.
Supervising where to crop may not solve this problem, as CodeV is explicitly rewarded for zooming into the target region but still has a high FNR.
If this hypothesis holds, we can improve faithfulness by specifying an appropriate training curriculum.
Since the model's accuracy without tools improves during training, we select training questions for which the correct zoomed evidence helps the current model answer correctly (\cref{sec:selection}).
Measuring this benefit requires annotated evidence regions, which is provided by our data synthesis pipeline (\cref{sec:data-synthesis}).

\section{Methodology}
\label{sec:method}

We train the policy on questions where zoomed evidence improves answer accuracy.
In this section, we first introduce our visual search data synthesis method (\cref{sec:data-synthesis}).
We then introduce our data selection method, which identifies the questions that the current policy answers more accurately with the ground-truth evidence crops than with the whole image alone (\cref{sec:selection}).
We also reward correct direct answers when the sampled trajectories indicate that the question can be answered without zooming (\cref{sec:reward}).

\subsection{Visual-search Data Synthesis} \label{sec:data-synthesis}

We aim to generate questions that are challenging to answer from the whole image alone without zooming.
As shown in~\cref{fig:method-overview}, the synthesis has three stages.
We first locate small objects in each high-resolution image, with image sources given in~\cref{sec:settings}.
A question generator then uses crops of these objects and their surroundings to write questions about object attributes and spatial relations.
Finally, we verify that each question refers to an identifiable target, has an answer supported by the image, and cannot be guessed from the question alone.
More details of data synthesis are provided in~\cref{app:data-construction} and~\cref{app:method-prompts}.

\paragraph{Object localisation.}
To identify objects that may be overlooked in the image, we divide each high-resolution image $I$ into smaller regions.
For each region, we use the question generator to list visible object types, then use SAM3~\citep{carion2025sam3} to obtain precise bounding boxes for the matched objects.
We map the bounding boxes from the smaller regions back to the coordinates of $I$, merge overlapping bounding boxes, and keep the top 32 objects based on SAM3's object confidence for each image.
For an object $o_i$, let $I_{o_i}$ denote the object crop and $v^+_{o_i}$ denote the ground-truth crop region that includes the object and its surroundings. 

\paragraph{Question generation.}
We construct two types of questions from the located objects: single-object questions and spatial-relation questions.

For generating a single-object question, we provide the generator with the object crop $I_{o_i}$, the object crop with its surroundings $v^+_{o_i}$, and the downscaled overview $I$ for broader context, where $v^+_{o_i}$ and $I$ help the generator describe the target's location and distinguish it from other objects in the image.
We predefine a list of visual search abilities: text recognition, colour recognition, object identification, counting, structure recognition, and material recognition.
The generator proposes a question $q$ targeting a specific ability, together with a draft answer $\tilde y$. The answer $\tilde y$ will be further checked or rewritten in the next step. 

For a spatial-relation question, the generator selects a second object $o_j$ from the saved detections in the surrounding region. It examines the object crops $I_{o_i}$ and $I_{o_j}$, the ground-truth crops $v^+_{o_i}$ and $v^+_{o_j}$, and the downscaled overview $I$ to propose a question and answer about a visible relationship, such as relative position, contact, or containment.

For both question types, we store the draft answer and the ground-truth crops of the question, $V^+=\{v^+_{o_i}\}$ for a single-object question or $V^+=\{v^+_{o_i},v^+_{o_j}\}$ for a relation question, which are used in verification and data selection.

\paragraph{Question verification.}
We check whether each question identifies the intended objects, whether the draft answer is determined when given the zoomed region, and whether the question can be guessed without seeing the image.

\emph{Target identification.}
Several objects may match a question's description, making its intended target ambiguous.
We check if the question is ambiguous using Set-of-Mark prompting~\citep{yang2023setofmark}.
Specifically, we mark SAM3-identified objects in the surrounding image region with numbered boxes and ask the question generator to select every object that matches the question's intended target. If the generator does not select exactly the intended target, we ask it to revise the question using the target's relative location or nearby landmarks, without leaking its absolute position. We then verify the revised question and reject it if the intended target still cannot be uniquely identified.

\emph{Answer verification.}
We sample multiple answers to $q$ using the whole image $I$, the object crops $I_{o}$, and the ground-truth crops $V^+$, without conditioning on the draft answer $\tilde y$.
The generator then reviews these answers alongside $\tilde y$ and proposes the final reference answer $y$. We discard the question if the generator judges the visual evidence insufficient to determine an answer.

\emph{Visual dependence.}
We ask the generator to answer $q$ without seeing the image, and reject the question if it answers correctly. This ensures that the question cannot be solved without visual grounding. Finally, each synthesized sample is then a tuple $(I, q, y, V^+)$.

\paragraph{Unanswerable questions.}
\label{sec:answerability}
To encourage the model to base its answers on visual evidence, we construct additional unanswerable examples by obscuring the regions needed to answer each question. Specifically, we obscure the region around the target object, either by replacing it with a random patch from the same image or by blurring it with Gaussian noise, to obtain an edited image $I^{\mathrm{hide}}$. The generator then judges whether the question is still answerable from $I^{\mathrm{hide}}$, and we keep the example only if the generator confirms that $I^{\mathrm{hide}}$ is insufficient to answer it.
For these examples, the model should state that the question is unanswerable rather than guessing.
To prevent the model from using the editing artefacts alone as a cue for unanswerability, we also build answerable counterparts by applying the same edits to randomly sampled regions that do not overlap the evidence region, yielding $I^{\mathrm{rand}}$. The generator checks these images as well, and we keep an example only if it remains answerable with the original answer unchanged. An unanswerable example is shown in~\cref{fig:curation-removal}.

\subsection{Adaptive data selection}
\label{sec:selection}
A question that initially requires zooming may become answerable from the full image alone as the model improves during training. We therefore measure the benefit of supplying the ground-truth crop before selecting an example for training. For each example $(I, q, y, V^+)$, we compare answering with and without the ground-truth crops $V^+$.

Let $\pi_t$ denote the policy at training step $t$. We sample answers from $\pi_t$ given either the whole image alone or the whole image together with $V^+$. In both conditions, the policy answers directly from the supplied images without calling tools. Because $V^+$ is provided rather than predicted by the policy, this comparison measures the benefit of the evidence itself, independent of the policy's ability to predict the targeted crop. The expected accuracies under $\pi_t$ are:
\begin{equation}
\begin{aligned}
&\mathrm{acc}_{\pi_t}(I,q)=\mathbb E_{\hat y\sim\pi_t(\cdot\mid I,q)}\left[\ind[\hat y=y]\right],\\
&\mathrm{acc}_{\pi_t}(I,V^+,q)=\mathbb E_{\hat y\sim\pi_t(\cdot\mid I,V^+,q)}\left[\ind[\hat y=y]\right].
\end{aligned}
\label{eq:selection}
\end{equation}
Here, $\ind[\hat y=y]$ indicates that the sampled answer is correct. We estimate both accuracies from sampled answers and select questions satisfying
\begin{equation}
\mathrm{acc}_{\pi_t}(I,V^+,q)-\mathrm{acc}_{\pi_t}(I,q)\geq\delta,
\label{eq:selection-criteria}
\end{equation}
where $\delta>0$ is the minimum accuracy improvement from the ground-truth crops over the whole image alone.
We calculate accuracies using the current policy model $\pi_t$, so selection adapts to the model's evolving capability during training.
During RL, the policy receives only the image $I$ and question $q$ and decides whether and where to inspect; $V^+$ is used for selection and is not supplied in the training trajectory. More details are given in~\cref{app:training}.

\begin{table*}[t]
\centering
\resizebox{0.88\textwidth}{!}{
\begin{tabular}{lrrrrrr}
\toprule
Model & V* & HR-Bench 4K & HR-Bench 8K & MME-RW-en & MME-RW-cn & ZoomBench \\
\midrule
Qwen2.5-VL-7B\textsuperscript{\dag}~\citep{bai2025qwen25vl} & $78.5$ & $71.6$ & $67.9$ & $60.8$ & $58.3$ & $42.5$ \\
PixelReasoner~\citep{wang2025pixelreasoner} & $84.3$ & $72.6$ & $66.1$ & $64.4$ & $60.5$ & $47.8$ \\
DeepEyes~\citep{zheng2025deepeyes} & $85.6$ & $75.1$ & $72.6$ & $64.1$ & $64.2$ & $46.6$ \\
Thyme~\citep{zhang2025thyme} & $82.2$ & $77.0$ & $72.0$ & $64.8$ & $65.0$ & $44.9$ \\
DeepEyesV2~\citep{hong2025deepeyesv2} & $81.8$ & $77.9$ & $73.8$ & $64.9$ & $64.8$ & $45.2$ \\
Mini o3~\citep{lai2025minio3} & $88.2$ & $77.5$ & $73.3$ & $65.5$ & $64.8$ & $50.3$ \\
Adaptive CoF~\citep{zhang2025adaptivecof} & \textbf{90.1} & $73.5$ & $67.6$ & -- & -- & -- \\
PyVision-Image~\citep{zhao2026pyvisionrl} & $88.7$ & $78.1$ & $74.3$ & -- & -- & -- \\
ZwZ-7B\textsuperscript{\dag}~\citep{wei2026zwz} & $88.5$ & $75.4$ & $73.3$ & \textbf{66.2} & \textbf{67.0} & $55.6$ \\
CodeV~\citep{you2025codev} & $84.8$ & $76.1$ & $71.3$ & $63.8$ & $63.2$ & $50.5$ \\
\midrule
\MethodName (Ours) & \textbf{90.1} & \textbf{79.5} & \textbf{77.3} & $65.9$ & \textbf{67.0} & \textbf{56.8} \\
\bottomrule
\end{tabular}
}
\caption{Answer accuracy (\%) on the perception-intensive benchmarks. Bold marks the highest score in each column. Results of Qwen2.5-VL-7B are taken from \citet{wei2026zwz}. $^\dag$ denotes models that do not crop or zoom into the image at inference time.}
\label{tab:main}
\end{table*}

\subsection{Adaptiveness reward for selection}
\label{sec:reward}
Because selection relies on the policy's tool-free accuracy, it can underestimate what the policy is able to answer from the whole image when the policy rarely answers directly and almost always invokes the tool.
To address this, we add an adaptiveness reward that favours correct tool-free answers on questions the policy can already answer reliably without tools. This makes the tool-free accuracy a more faithful estimate of what the policy can answer without zooming.
  
For each image--question pair, we sample a group $\mathcal G=\{\tau_i\}_{i=1}^{G}$ of trajectories (\cref{eq:trajectory}). Let $r_i^{\mathrm{acc}}$ be 1 if $\tau_i$ answers correctly and 0 otherwise, and let $u_i=\ind[K_i>0]$ indicate whether $\tau_i$ calls the tool, where $K_i$ is its number of tool calls (\cref{eq:trajectory}). We define group accuracy as $a_{\mathcal G}=\frac{1}{G}\sum_{j=1}^{G}r_j^{\mathrm{acc}}$ and tool-free group accuracy as $d_{\mathcal G}=\frac{1}{G}\sum_{j=1}^{G}r_j^{\mathrm{acc}}(1-u_j)$. The latter is the proportion of all trajectories in the group that answer correctly without tools. Then we define the adaptiveness reward:
\begin{equation}
r_i^{\mathrm{adapt}}=\begin{cases}
r_i^{\mathrm{acc}}(1-u_i), & a_{\mathcal G}\geq\alpha,\ d_{\mathcal G}\geq\rho,\\
0, & \text{otherwise},
\end{cases}
\label{eq:adaptiveness}
\end{equation}
where $\alpha$ and $\rho$ are the minimum group accuracy and tool-free group accuracy. Only correct trajectories without tool calls receive the adaptiveness reward, and only when the group satisfies both thresholds.
The final reward consists of answer correctness reward $r_\mathrm{acc}$,  adaptiveness reward  $r_\mathrm{adapt}$, and format correctness reward $r_\mathrm{fmt}$:
\begin{equation}
r_i=w_{\mathrm{acc}}r_i^{\mathrm{acc}}+w_{\mathrm{adapt}}r_i^{\mathrm{adapt}} + w_{\mathrm{fmt}}r_i^{\mathrm{fmt}},
\label{eq:reward}
\end{equation}
where $w_{\mathrm{acc}}$, $w_{\mathrm{adapt}}$, and $w_{\mathrm{fmt}}$ control their relative weights. We optimise the policy with GRPO~\citep{shao2024deepseekmath}. 

The adaptiveness reward and selection work together: the reward keeps the policy answering without tools when it can, so that its accuracy without tools reflects what it can see in the whole image, and selection uses this accuracy to select questions where the ground-truth crops can help.
They are intended to adapt the training data to the model's evolving capability during training.

\section{Experiments} \label{sec:experiments}

\subsection{Experimental Settings} \label{sec:settings}

\begin{table}[t]
\centering
\resizebox{\linewidth}{!}{%
\begin{tabular}{lrrr}
\toprule
Model & V* & HR-Bench 8K & ZoomBench \\
\midrule
\MethodName & \textbf{90.1} & \textbf{77.3} & \textbf{56.8} \\
\quad w/o selection & $86.4$ & $71.5$ & $48.2$ \\
\quad w/o adapt. reward & \textbf{90.1} & $74.8$ & $56.3$ \\
\quad w/o sel. \& adapt. reward & \textbf{90.1} & $71.9$ & $55.5$ \\
\quad w/o synthesised data & $85.5$ & $69.0$ & $49.5$ \\
\quad w/o tool training & $86.9$ & $70.6$ & $51.0$ \\
\quad w/o unanswerable data & $89.4$ & $74.6$ & $55.5$ \\
\quad w/o SFT & $88.0$ & $73.3$ & $52.9$ \\
\bottomrule
\end{tabular}
}
\caption{Answer accuracy (\%) of the ablation models. Adapt. reward denotes the adaptiveness reward.}
\label{tab:ablation}
\end{table}

\paragraph{Training data.}
For RL training, we use our data synthesis workflow (\cref{sec:data-synthesis}) to construct 13K visual-search questions on high-resolution images from CC12M~\citep{changpinyo2021cc12m}, SA-1B~\citep{kirillov2023segment}, InfographicVQA~\citep{mathew2022infographicvqa}, Wikimedia Commons~\citep{wikimediaCommons}, VisDrone~\citep{zhu2021visdrone}, TinyPerson~\citep{yu2020tinyperson}, Unsplash~\citep{unsplash}, DOTA~\citep{xia2018dota}, DocVQA~\citep{mathew2021docvqa}, SKU-110K~\citep{goldman2019sku110k}, and xView~\citep{lam2018xview}.
Of these questions, $67.5\%$ are single-object questions, $25.0\%$ are spatial-relation questions, and $7.5\%$ are unanswerable questions, and \cref{tab:data-ability} presents their distribution across abilities.
Before RL, we apply SFT for a cold start so that the policy learns the tool-call format and how to reason about a returned crop.
We construct 1K SFT examples from questions produced by our data synthesis method.
Each example shows one of four behaviours: answering directly from the whole image, zooming into the ground-truth crop and then answering, recovering after a first zoom that misses the target, or searching several regions before concluding that the question is unanswerable.
More details of the RL and SFT data are provided in \cref{app:data-construction,app:sft}.

\suppressfloats[t]
\begin{figure}[t]
\centering
\includegraphics[width=\linewidth]{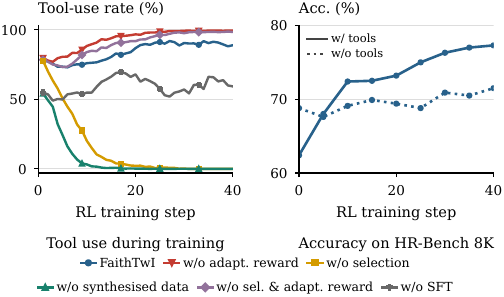}
\caption{Tool use rate and answer accuracy during training. Left: percentage of training responses containing a tool call. Right: HR-Bench 8K accuracy of \MethodName with tools (solid) and without tools during inference time (dotted).}
\label{fig:ablation-tool-dynamics}
\end{figure}

\begin{table}[t]
\centering
\resizebox{0.88\linewidth}{!}{
\begin{tabular}{lrrr}
\toprule
Model & FNR$\downarrow$ & FPR$\downarrow$ & Acc.$\uparrow$ \\
\midrule
Full method & $8.7$ & $2.2$ & $94.5$ \\
\quad w/o adapt. reward & $9.8$ & $4.9$ & $92.6$ \\
\quad w/o selection \& adapt. reward & $8.9$ & $7.4$ & $91.9$ \\
\quad w/o unanswerable data & $13.3$ & $3.7$ & $91.5$ \\
\quad w/o SFT & $0.0$ & $96.3$ & $51.9$ \\
\bottomrule
\end{tabular}
}
\caption{Target-recognition accuracy and error rates after replacing the first returned crop on V* (\%).}
\label{tab:crop-ablation}
\end{table}

\paragraph{Evaluation benchmarks and baselines.}
We evaluate visual search and fine-grained perception on six benchmarks: V*~\citep{gao2024vstar}, HR-Bench 4K and 8K~\citep{ding2024hrbench}, the English and Chinese splits of MME-RealWorld~\citep{zhang2024mmerealworld}, and ZoomBench~\citep{wei2026zwz}.
We further assess faithfulness on four multimodal hallucination benchmarks: RePOPE~\citep{DBLP:conf/emnlp/LiDZWZW23,DBLP:journals/corr/abs-2504-15707}, AMBER~\citep{DBLP:journals/corr/abs-2311-07397}, HaloQuest~\citep{DBLP:conf/eccv/WangBYLLG24}, and VizWiz~\citep{DBLP:conf/cvpr/Gurari0SGLGLB18}. In the main paper, we compare baselines initialised from Qwen2.5-VL-7B: DeepEyes~\citep{zheng2025deepeyes}, PixelReasoner~\citep{wang2025pixelreasoner}, Thyme~\citep{zhang2025thyme}, DeepEyesV2~\citep{hong2025deepeyesv2}, Mini-o3~\citep{lai2025minio3}, Adaptive CoF~\citep{zhang2025adaptivecof}, PyVision-Image~\citep{zhao2026pyvisionrl}, CodeV~\citep{you2025codev}, and ZwZ~\citep{wei2026zwz}.

\paragraph{Implementation details.}
We use Qwen2.5-VL-7B~\citep{bai2025qwen25vl} as the main base model because most existing TwI models are built on it, allowing us to analyse and compare the faithfulness and performance more effectively.
Results based on Qwen3-VL-4B~\citep{qwen3vl} and Qwen3.5-4B~\citep{qwen2026qwen35} are reported in \cref{app:qwen3vl}. SFT runs for one epoch (32 steps) with a batch size of 32 and a learning rate of $10^{-5}$.
We then train with GRPO for 40 steps, with a learning rate of $10^{-6}$ and 384 questions per step, sampling 12 trajectories per question.
During both training and inference, we allow up to eight interaction turns, i.e., $K=8$ in~\cref{eq:trajectory}. More training and inference details are provided in \cref{app:training}.

\subsection{Main Experimental Results}
\label{sec:main-results}

\suppressfloats[t]
\begin{table}[t]
\centering
\resizebox{\linewidth}{!}{%
\begin{tabular}{lrrrr}
\toprule
Model & RePOPE & AMBER & HaloQuest & VizWiz \\
\midrule
PixelReasoner & \textbf{93.6} & \textbf{87.9} & $54.4$ & $49.1$ \\
DeepEyes & $88.3$ & $85.2$ & $47.5$ & $46.1$ \\
Thyme & $89.4$ & $85.2$ & $34.5$ & $44.3$ \\
DeepEyesV2 & $43.7$ & $38.2$ & $23.4$ & $45.8$ \\
Mini o3 & $73.7$ & $75.6$ & $26.2$ & $44.4$ \\
CodeV & $91.1$ & $84.7$ & $62.2$ & $54.9$ \\
\midrule
\MethodName & \textbf{93.6} & $87.3$ & \textbf{70.7} & \textbf{57.1} \\
\quad w/o adapt. reward & $92.7$ & $84.2$ & $56.7$ & $50.4$ \\
\quad w/o sel. \& adapt. reward & $91.9$ & $84.5$ & $62.0$ & $50.6$ \\
\quad w/o unanswerable data & $79.7$ & $73.7$ & $31.7$ & $40.3$ \\
\quad w/o SFT & \textbf{93.6} & $87.7$ & $34.2$ & $45.0$ \\
\bottomrule
\end{tabular}
}
\caption{Faithfulness on multimodal hallucination benchmarks. Bold marks the highest score in each column.}
\label{tab:faith-benchmark-ablation}

\end{table}

\paragraph{Visual search evaluation results.}
As shown in~\cref{tab:main}, \MethodName achieves the highest accuracy compared to strong baselines on V*, HR-Bench, and ZoomBench. On V*, it reaches $90.1\%$, matching Adaptive CoF and at least $1.4\%$ higher than the other baselines.
HR-Bench and ZoomBench further evaluate fine-detail perception in high-resolution images. Our model achieves $79.5\%$ on HR-Bench 4K, $77.3\%$ on HR-Bench 8K, and $56.8\%$ on ZoomBench, improving over the strongest baselines by $1.4\%$, $3.0\%$, and $1.2\%$, respectively.
On MME-RealWorld, \MethodName achieves comparable accuracy to ZwZ-7B.
The largest gain occurs on HR-Bench 8K, suggesting that the zooming is particularly useful when process higher-resolution images.

\paragraph{Faithfulness evaluation results.}
The target recognition audit in~\cref{sec:diagnosis} shows that \MethodName achieves $94.5\%$ target-recognition accuracy, with $8.7\%$ FNR and $2.2\%$ FPR.
To investigate whether faithful target recognition also extends to answer accuracy, we further evaluate on the multimodal hallucination benchmarks RePOPE, AMBER, HaloQuest, and VizWiz.
As shown in~\cref{tab:faith-benchmark-ablation}, our model outperforms DeepEyes and CodeV on all four benchmarks and achieves the highest scores among the listed models on HaloQuest and VizWiz, exceeding CodeV by $8.5\%$ and $2.2\%$, respectively.
PixelReasoner achieves comparable accuracy compared to \MethodName on RePOPE and AMBER but much lower on HaloQuest and VizWiz.

\subsection{Analysis and Discussion}
\label{sec:ablations}

We examine how data selection, the adaptiveness reward, synthesised questions, and tool access during training affect accuracy and visual-evidence use. Starting from the same SFT checkpoint, we train models without selection (w/o selection), without the adaptiveness reward (w/o adapt. reward), and without either component (w/o selection \& adapt. reward). We also replace the synthesised training dataset with DeepEyes questions (w/o synthesised data) or remove tool access during RL (w/o tool training). More implementation details are described in~\cref{app:result-protocols}.

\paragraph{Selection and the adaptiveness reward jointly shape tool use.}
As shown in~\cref{fig:ablation-tool-dynamics} (left), without data selection, tool use falls below $1\%$ by step 24, and without synthesized data, it falls below $1\%$ by step 13. In contrast, removing the adaptiveness reward (w/o adapt. reward) or both selection and the reward (w/o sel. \& adapt. reward) keeps tool use at $99.4\%$ and $98.3\%$ at the end of training, compared with $88.8\%$ for the full method.
This indicates that selection helps maintain tool use when the adaptiveness reward encourages correct direct answers, while the adaptiveness reward slightly lowers the tool-use rate.
The collapse of tool use without selection hurts fine-grained perception. As shown in~\cref{tab:ablation}, removing selection alone lowers accuracy on HR-Bench 8K to $71.5\%$, compared with $77.3\%$ for the full method, because the model no longer invokes the tool to inspect fine-grained regions of the high-resolution image. Removing both selection and the reward gives a similar $71.9\%$.

\paragraph{Selection and the adaptiveness reward jointly improve faithfulness.}
We compare faithfulness among the models that continue to use tools: the full method, the model without the adaptiveness reward, and the model without both selection and the reward.
As shown in~\cref{tab:crop-ablation}, the full method achieves a crop-recognition accuracy of $94.5\%$, compared with $92.6\%$ without the adaptiveness reward and $91.9\%$ without both components. \Cref{tab:faith-benchmark-ablation} shows that the full method obtains the highest score on all four multimodal hallucination benchmarks. For example, on HaloQuest, it reaches $70.7\%$, compared with $56.7\%$ without the reward and $62.0\%$ without both components. Selection without the adaptiveness reward does not consistently improve these scores. These results indicate that adaptive data selection and the adaptiveness reward jointly improve faithfulness.

\paragraph{Tool use improves answer accuracy rather than being decorative.}
We examine tool use in two settings: 1) we evaluate \MethodName with and without tool access at inference time (\cref{fig:ablation-tool-dynamics}, right); 2) we compare \MethodName with the model without tools during RL training (w/o tool training) (\cref{tab:ablation}). The first comparison measures the benefit of tools at inference, and the second comparison isolates the benefit of training on our dataset without the tool.
At inference, tool access initially lowers HR-Bench 8K accuracy from $68.8\%$ to $62.4\%$, but the final model gives $77.3\%$ with tools and $71.5\%$ without them. Thus, the trained model benefits from using tools beyond what it can answer directly. The model trained and evaluated without tools reaches $70.6\%$ on HR-Bench 8K and $51.0\%$ on ZoomBench, below the full method's $77.3\%$ and $56.8\%$.
These two results demonstrate that tool use improves accuracy rather than being decorative.

\paragraph{Effectiveness of the synthesised training data.}
Replacing our synthesised training dataset with DeepEyes training data (w/o synthesised data) leads to almost no tool use during training and reduces accuracy to $69.0\%$ on HR-Bench 8K and $49.5\%$ on ZoomBench, compared with $77.3\%$ and $56.8\%$ for the full method. It also omits data selection because the data lack the evidence-region annotations needed to estimate the benefit of zoomed evidence specified in~\cref{eq:selection}. Thus, it combines synthesised questions with evidence regions that enable selection, rather than isolating question synthesis alone; instead, without selection ablation (w/o selection) shows the effectiveness of data selection.

\paragraph{Lightweight SFT is necessary for faithful tool use.}
As shown in~\cref{tab:crop-ablation}, FPR increases to $96.3\%$ without SFT, compared with $2.2\%$ for the full method.
Removing SFT also reduces HaloQuest accuracy from $70.7\%$ to $34.2\%$ and VizWiz accuracy from $57.1\%$ to $45.0\%$ (\cref{tab:faith-benchmark-ablation}).
These results suggest that SFT provides an important starting point; starting from the SFT model, data selection and the adaptiveness reward improve accuracy and faithfulness effectively.
CodeV~\citep{you2025codev} uses 333K SFT examples, whereas \MethodName uses only 1K examples for one epoch, showing that lightweight SFT is necessary and sufficient.

\begin{figure}[t]
\centering
\includegraphics[width=0.88\linewidth]{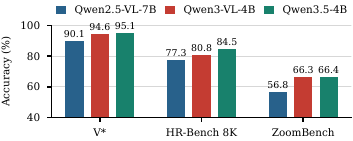}
\caption{Answer accuracy (\%) of \MethodName trained from three base models: Qwen2.5-VL-7B, Qwen3-VL-4B, and Qwen3.5-4B.}
\label{fig:other-bases}
\end{figure}

\paragraph{Results with more base models.}
To test whether our method transfers to other base models, we also apply it to Qwen3-VL-4B-Thinking~\citep{bai2025qwen3vl} and Qwen3.5-4B~\citep{qwen2026qwen35}, with the same training settings. As shown in~\cref{fig:other-bases}, \MethodName reaches $80.8\%$ on HR-Bench 8K and $66.3\%$ on ZoomBench with Qwen3-VL-4B, $84.5\%$ and $66.4\%$ with Qwen3.5-4B. Comparisons with other models based on Qwen3-VL-4B and Qwen3.5-4B are presented in~\cref{app:qwen3vl}.

\section{Related Work}

\paragraph{Thinking with images.}
VLMs often miss small objects and fine spatial details in high-resolution, information-dense images. Thinking with images addresses this by letting the model crop and zoom into regions during reasoning, a paradigm popularised by OpenAI o3~\citep{openai2025o3}. PixelReasoner~\citep{wang2025pixelreasoner}, DeepEyes~\citep{zheng2025deepeyes}, Chain-of-Focus~\citep{zhang2025adaptivecof}, ViGoRL~\citep{sarch2025vigorl}, and Mini-o3~\citep{lai2025minio3} train VLMs to use crop-and-zoom tools with RL, and Thyme~\citep{zhang2025thyme}, CodeV~\citep{you2025codev}, PyVision-RL~\citep{zhao2026pyvisionrl}, and DeepEyesV2~\citep{hong2025deepeyesv2} extend the tools to code execution and search. Another line of work removes the tool at inference time by distilling from zoomed images~\citep{wei2026zwz,yuan2026visionopd}. 
Unlike distillation, which hides zooming at inference, explicit zooming shows which region an answer is based on. This is useful only if the model faithfully uses the zoomed region, which is what we study in this work.

\paragraph{Faithfulness of visual tool use.}
Chain-of-thought explanations can misrepresent what drives a prediction~\citep{turpin2023language,lanham2023measuring}, and recent work finds a similar gap for visual tools: most of the gains from tool-use RL come from intrinsic capability rather than from the returned images~\citep{ma2026med,guo2026benefit,shao2026textcall}. Other studies test whether the visual evidence changes the answer~\citep{wang2026illusion,zhang2026insensitivity,li2026viebench}, whereas we analyse how models respond to correct and wrong crops.
Some prior works supervise reasoning process to improve faithful visual tool use~\citep{you2025codev,wang2025virl,yang2026mapo}. Other work filters data by tool benefit before training~\citep{zheng2025deepeyes,hong2025deepeyesv2,wang2026adatoolerv,mao2026toolvision,xiang2026openvistool}. We instead improve faithfulness through data synthesis and selection with outcome-level rewards, estimating the benefit of zoomed evidence during training.

\section{Conclusion}

Through a target recognition analysis, we find that TwI models are often unfaithful to the images returned by the tool. 
We trace this behaviour to training data on which zooming does not improve accuracy. To address this, we synthesise challenging questions, select training data according to the policy's evolving capability, and reward correct tool-free trajectories. Experimental results show that \MethodName improves both accuracy and faithfulness.

\section*{LLM Usage Statement}
We used LLM-based agents, including Claude Code and Codex, to help review our code and to check the writing for grammar, wording, and clarity, in a similar way to grammar-checking tools. As part of our method, Qwen3.8-27B generates and verifies the training questions, and Qwen-Flash-Next serves as the judge.

\subsection*{Acknowledgments}
Yu Zhao was partly supported by the UKRI Centre for Doctoral Training in Natural Language Processing, funded by UK Research and Innovation (grant EP/S022481/1) and the University of Edinburgh, School of Informatics.
Rohit Saxena was supported by the Engineering and Physical Sciences Research Council (EPSRC) through the AI Hub in Generative Models (grant number EP/Y028805/1).
Pasquale Minervini was partially funded by ELIAI (The Edinburgh Laboratory for Integrated Artificial Intelligence), EPSRC (grant no.\ EP/W002876/1), and a donation from Accenture LLP.

{
    \small
    \bibliographystyle{ieeenat_fullname}
    \bibliography{custom}
}

\clearpage
\appendix

\section{Evidence Recognition Experiment}
\label{app:section3-protocol}

\paragraph{Setup.}
We use V*~\citep{gao2024vstar} for the analysis because it provies evidence region annotation. We run V* three times and analyse the first crop of each trajectory in two settings.
\begin{itemize}[leftmargin=*,nosep]
\item \emph{Controlled.} This is what we report in the main paper. We keep each trajectory up to its first tool call and replace the returned crop. The replacement either shows the target, as a ground-truth crop centred on the annotated target box, or does not, as a random crop, a uniform grey image, or a uniform black image.
\item \emph{Real.} The model's own first crop is returned unchanged. We count it as showing the target if it covers at least $90\%$ of an annotated target box and as missing the target if it covers less than $10\%$, and leave out the remaining crops.
\end{itemize}
The main paper reports the controlled setting with ground-truth and random crops. Whether the target is visible is known by construction, so no crop needs to be labelled, and crops with and without the target are balanced.

\paragraph{Metrics.}
For each response, the judge decides whether the model claims that the target is visible. FNR is the percentage of crops showing the target for which the model claims that the target is not visible, and FPR is the percentage of crops without the target for which it claims that the target is visible. Accuracy is computed on ground-truth and random crops, as in \cref{tab:evidence-use}.

\paragraph{Results.}
\Cref{tab:section3-counts} reports both settings. DeepEyes and PixelReasoner claim to see the target on most grey and black images, whereas CodeV and \MethodName rarely do. In the real setting, most crops show the target; for \MethodName, only 22 crops miss the target, so its real-crop FPR rests on few crops. These scores describe the first response after a crop, not the effect of the crop on the final answer.

\begin{table*}[t]
\centering\small
\setlength{\tabcolsep}{6pt}
\begin{tabular}{lrrrrrrr}
\toprule
 & \multicolumn{2}{c}{Real crops} & \multicolumn{5}{c}{Controlled crops} \\
\cmidrule(lr){2-3}\cmidrule(lr){4-8}
Model & FNR & FPR & FNR (ground truth) & FPR (random) & FPR (grey) & FPR (black) & Acc. \\
\midrule
DeepEyes & $0.0$ & $94.1$ & $1.6$ & $98.8$ & $97.6$ & $98.0$ & $50.0$ \\
PixelReasoner & $2.8$ & $91.9$ & $2.3$ & $56.9$ & $95.3$ & $95.1$ & $70.8$ \\
CodeV & $84.1$ & $10.7$ & $87.9$ & $0.7$ & $0.5$ & $0.0$ & $55.2$ \\
\MethodName & $6.1$ & $18.2$ & $8.7$ & $2.2$ & $0.7$ & $0.0$ & $94.5$ \\
\bottomrule
\end{tabular}
\caption{First-crop target recognition on V* (\%). Real crops are the models' own first crops. Controlled crops replace them with a ground-truth crop, which shows the target, or with a random crop, a grey image, or a black image, which do not. FNR is measured on ground-truth crops and FPR on each type of crop without the target; accuracy is measured on ground-truth and random crops.}
\label{tab:section3-counts}
\end{table*}

\begin{table*}[t]
\centering\small
\setlength{\tabcolsep}{5pt}
\begin{tabular}{lrrrrrrrrr}
\toprule
 & \multicolumn{3}{c}{V*} & \multicolumn{3}{c}{HR-Bench 4K} & \multicolumn{3}{c}{HR-Bench 8K} \\
\cmidrule(lr){2-4}\cmidrule(lr){5-7}\cmidrule(lr){8-10}
Model & With & Without & Tool use & With & Without & Tool use & With & Without & Tool use \\
\midrule
DeepEyes & $81.8$ & $83.8$ & $100.0$ & $74.5$ & $73.9$ & $100.0$ & $70.9$ & $69.5$ & $99.9$ \\
PixelReasoner & $82.5$ & $79.9$ & $75.9$ & $72.9$ & $74.6$ & $51.8$ & $65.4$ & $66.5$ & $41.4$ \\
CodeV & $80.6$ & $77.7$ & $96.3$ & $75.0$ & $72.8$ & $94.2$ & $69.4$ & $68.2$ & $94.2$ \\
\bottomrule
\end{tabular}
\caption{Accuracy with and without tools, and tool-use rate (\%), evaluated by us with each model's official settings. Without tools, the prompts still ask the model to reason before answering. Tool use is the percentage of questions on which the model calls the tool. V* is averaged over three runs, and HR-Bench uses one run.}
\label{tab:observation-tool-accuracy}
\end{table*}

\section{DeepEyes Replication}
\label{app:deepeyes}
\label{app:observation-analysis}

We replicate the RL training of DeepEyes~\citep{zheng2025deepeyes} with its released code and official settings: GRPO from Qwen2.5-VL-7B-Instruct on the 47K DeepEyes training questions, with 256 questions and 16 rollouts per step for 80 steps, using our judge for the reward. \Cref{fig:observation-dynamics} evaluates its checkpoints every ten steps with the official DeepEyes evaluation settings: the DeepEyes prompt, at most 16,384 visual tokens for the image, greedy decoding, and at most five turns. Without tools, the tool is removed and the model answers in one turn. After 80 steps, the replicated model obtains $85.5\%$, $71.5\%$, and $69.9\%$ on V*, HR-Bench 4K, and HR-Bench 8K with tools, and $83.3\%$, $73.5\%$, and $68.9\%$ without tools. All other DeepEyes results use the released checkpoint, and the V* scores of DeepEyes and DeepEyesV2 in \cref{tab:main} are those reported by \citet{hong2025deepeyesv2}.

\section{Training Dataset Construction}
\label{app:data-construction}

\paragraph{Image sources.}
\Cref{sec:settings} lists the image sources. Most questions use images from CC12M~\citep{changpinyo2021cc12m} ($52.3\%$) and SA-1B~\citep{kirillov2023segment} ($31.5\%$). Images that appear in our validation or test sets are excluded from training.

\paragraph{Object localisation.}
We use images whose longer side is at least 1,000 pixels. We split each image into tiles of $1024\times1024$ pixels, with a $25\%$ overlap between neighbouring tiles. For each tile, the question generator lists the visible object types, and SAM3~\citep{carion2025sam3} localises objects of these types in the tile (\cref{sec:data-synthesis}). We map the boxes back to the original image, keep those with a SAM3 confidence of at least 0.75, and merge duplicate boxes from overlapping tiles. We keep objects whose box is at most $64\times64$ pixels and does not touch the image boundary, since a cut-off object may not determine the answer. Each ground-truth crop contains the object box and its surroundings; near the image boundary, the crop window is shifted inward rather than clipped, so that it keeps its full size.

\paragraph{Spatial-relation questions.}
For a spatial-relation question, we take a $1920\times1080$-pixel window centred on the box of the first object, shifted inward at the image boundary and limited to the image size. The question generator chooses the second object among the localised objects whose boxes lie entirely inside this window, and both objects satisfy the same size limit. The question refers to both objects only by their descriptions.

\paragraph{Unanswerable questions.}
\label{app:variants}
We construct unanswerable questions from verified open-ended synthesised questions by obscuring their evidence region: we blur it, paste a patch from the same image over it, or fill it with grey. The whole image and every crop are then taken from the edited image, so the evidence cannot be recovered by zooming. We keep an edited question as unanswerable, with the answer \abstain, only if it passes two checks: the question generator judges that the answer can no longer be determined from the edited image, and the current policy cannot answer correctly to the original answer given the edited image and the edited evidence crop. We also apply the same kinds of edits to a region away from the evidence of other questions, which keep their original answers. Unanswerable questions are not filtered by the selection criterion in \cref{eq:selection}.

\paragraph{Training batch data proportions.}
Each batch contains 384 questions, including 16 unanswerable questions and 16 answerable questions whose images are edited away from the evidence. The target ability proportions are 20\% each for colour recognition and spatial relations, 15\% for text recognition, 10\% each for charts and counting, 8\% for entity recognition, and 7\% for the remaining abilities. Within each ability, no single answer accounts for more than 10\% of the selected questions; for example, without this limit, ``white'' appears too often as the answer to colour-recognition questions. The actual proportions depend on the available questions and on selection with the current policy.

\begin{table}[!ht]
\centering\small
\begin{tabular}{lrr}
\toprule
Ability & Questions & Proportion (\%) \\
\midrule
Spatial relation & $3{,}655$ & $28.1$ \\
Colour recognition & $3{,}382$ & $26.0$ \\
Text recognition & $2{,}054$ & $15.8$ \\
Structure recognition & $1{,}650$ & $12.7$ \\
Entity recognition & $1{,}176$ & $9.0$ \\
Counting & $560$ & $4.3$ \\
Material recognition & $524$ & $4.0$ \\
\bottomrule
\end{tabular}
\caption{Ability distribution of the synthesised training questions.}
\label{tab:data-ability}
\end{table}

\section{Training Details}
\label{app:training}
\label{app:sft}

We use Qwen3.8-27B~\citep{qwen2026qwen38_27b} as the question generator and Qwen-Flash-Next~\citep{qwen2026qwen38flashnext} as the judge that compares answers with references. We resize the image to at most 10240 visual tokens and each crop to at most 2048 visual tokens. We use SGLang~\citep{zheng2024sglang} for inference with greedy decoding. The ablation models use the same settings, except where removing a component requires a change.

\paragraph{SFT data.}
Before RL, SFT teaches the policy the tool-call format and how to reason about each returned crop. Each SFT example is a complete multi-turn conversation built from a synthesised question with an answer and its evidence boxes (\cref{sec:data-synthesis}). The examples cover four behaviours:
\begin{itemize}[leftmargin=*,nosep]
\item \emph{Direct answer.} The model answers from the whole image without zooming, either with the tool available or with no tool offered.
\item \emph{Zoom and answer.} The model zooms into the evidence region, or into the regions of both objects for a spatial-relation question, and answers.
\item \emph{Recovery.} The first zoom misses the target, for example by landing next to it or on a featureless region. The model notices that the target is not in the crop, zooms into the correct region, and answers.
\item \emph{Unanswerable question.} The evidence needed to answer is not in the image. The model inspects two or three regions, or the whole image when no tool is offered, before stating that the answer cannot be determined.
\end{itemize}
To decide between a direct answer and zooming, we ask the base Qwen2.5-VL-7B-Instruct each answerable question four times with the whole image. We use the questions it answers correctly three or four times to build direct-answer examples, and the questions it answers correctly at most once to build zooming examples.

\paragraph{Synthesise SFT examples.}
We describe how we synthesise each category of the example in the following.
In a zooming example, the assistant's zoom call requests the ground-truth crop of the question; in a recovery example, the first call requests a wrong region, next to the target or on a featureless area. We run the zoom tool on these calls, so every returned crop is a real crop of the image. The judge model then writes the assistant's reasoning around these crops, given three hand-checked examples of the same behaviour as references. After each crop, the reasoning first describes what the crop shows, then states whether the target is fully, partly, or not visible, and finally answers or chooses another region. The judge model checks each reply against its crop; a reply that fails the check is rewritten, and the example is dropped if the rewrite also fails. We also check samples of each behaviour by hand against their images.

\paragraph{SFT settings.}
For an example with tokens $x_j$, the SFT objective is
\begin{equation}
\mathcal{L}_{\mathrm{SFT}}=-\sum_j m_j\log\pi_\theta(x_j\mid x_{<j}),
\end{equation}
where $m_j=1$ for tokens of supervised assistant turns and $m_j=0$ for the prompt, the tool outputs, and, in recovery examples, the turns requesting an unsuccessful zoom. The unsuccessful zooms and their crops thus stay in the context without being supervised, so the policy learns to recognise and correct a missed zoom without learning to make the miss. SFT uses a batch size of 32, a learning rate of $10^{-5}$, and one epoch with 6 warm-up steps. We freeze the vision encoder and projector during SFT.

\paragraph{RL settings.}
We sample $G=12$ trajectories for each of $B=384$ questions, yielding 4,608 trajectories per step. GRPO compares rewards within each question group. For the rewards $R_1,\dots,R_G$ of a group, the advantage of trajectory $i$ is
\begin{equation}
 A_i=\frac{R_i-\mathrm{Mean}(\{R_1,R_2,\dots,R_G\})}{\mathrm{Std}(\{R_1,R_2,\dots,R_G\})+\epsilon},
\end{equation}
where $\epsilon>0$ avoids division by zero. The advantage applies to the trajectory's generated policy tokens; tool-returned images and text are conditioning context. A clipped policy-ratio objective limits each update relative to the rollout policy. Groups with equal rewards provide no relative preference among their trajectories. Training uses two policy mini-batches of 192 questions, a learning rate of $10^{-6}$, lower and upper clipping parameters of 0.20 and 0.28, and token-mean loss aggregation. Neither a KL regularizer nor an entropy reward is applied. The vision encoder and projector are updated during RL. Rollouts use temperature 1.0, at most eight interaction turns, and prompt/response limits of 24,576/20,480 tokens. The whole image receives up to 10,240 visual tokens and each crop 256--1,280. Trajectories exceeding the interaction limit are excluded from the policy loss.

\begin{figure}[t]
\centering
\includegraphics[width=\linewidth]{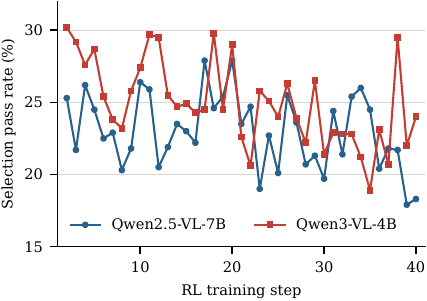}
\caption{Selection pass rate of the synthesised questions at each RL step for \MethodName, i.e., the percentage of checked questions that are used for training.}
\label{fig:selection-rate}
\end{figure}

\paragraph{Adaptiveness reward.}
We estimate both accuracies in \cref{eq:adaptiveness} from the $G=12$ trajectories sampled for each question. With $\alpha=0.5$ and $\rho=0.25$, the reward thus requires at least six correct answers overall and at least three correct answers without tools. We use $w_{\mathrm{acc}}=0.9$ and $w_{\mathrm{adapt}}=0.2$. The adaptiveness reward is applied only to answerable questions; for unanswerable questions, we set $r_i^{\mathrm{adapt}}=0$, including when the model correctly states that the question is unanswerable without using a tool.

\paragraph{Ablation configurations.}
\label{app:result-protocols}
The model without selection is trained on the same dataset with the same source proportions. The model without synthesised data uses the 47K DeepEyes questions with their original proportions. The model without tool training answers in one turn during RL and does not use tools during inference. All ablation models start from the same SFT model, except the model without SFT, which starts RL directly from Qwen2.5-VL-7B-Instruct.

\begin{table*}[t]
\centering
\resizebox{0.91\textwidth}{!}{
\begin{tabular}{lrrrrrr}
\toprule
Model & V* & HR-Bench 4K & HR-Bench 8K & MME-RW-en & MME-RW-cn & ZoomBench \\
\midrule
\MethodName & \textbf{90.1} & \textbf{79.5} & \textbf{77.3} & \textbf{65.9} & $67.0$ & \textbf{56.8} \\
\quad w/o selection & $86.4$ & $75.1$ & $71.5$ & $65.7$ & $65.6$ & $48.2$ \\
\quad w/o adapt. reward & \textbf{90.1} & $78.8$ & $74.8$ & $63.5$ & $64.9$ & $56.3$ \\
\quad w/o selection \& adapt. reward & \textbf{90.1} & $77.6$ & $71.9$ & $62.9$ & $63.8$ & $55.5$ \\
\quad w/o synthesised data & $85.5$ & $75.6$ & $69.0$ & $65.1$ & $66.4$ & $49.5$ \\
\quad w/o tool training & $86.9$ & $76.1$ & $70.6$ & $64.9$ & $65.3$ & $51.0$ \\
\quad w/o unanswerable data & $89.4$ & \textbf{79.5} & $74.6$ & $64.2$ & $64.8$ & $55.5$ \\
\quad w/o SFT & $88.0$ & $75.9$ & $73.3$ & $65.4$ & \textbf{67.1} & $52.9$ \\
\bottomrule
\end{tabular}
}
\caption{Answer accuracy (\%) of the ablation models on all six perception-intensive benchmarks. Bold marks the highest score in each column. Adapt. reward denotes the adaptiveness reward.}
\label{tab:ablation-full}
\end{table*}

\begin{table*}[t]
\centering
\resizebox{0.91\textwidth}{!}{
\begin{tabular}{lrrrrrr}
\toprule
Model & V* & HR-Bench 4K & HR-Bench 8K & MME-RW-en & MME-RW-cn & ZoomBench \\
\midrule
\multicolumn{7}{l}{\textit{Based on Qwen3-VL-4B}} \\
\midrule
Qwen3-VL-4B-Instruct\textsuperscript{\dag}~\citep{bai2025qwen3vl} & $80.1$ & $78.3$ & $72.9$ & $63.5$ & $63.6$ & $40.2$ \\
Qwen3-VL-4B-Thinking\textsuperscript{\dag}~\citep{bai2025qwen3vl} & $79.6$ & $74.5$ & $66.9$ & $60.3$ & $58.8$ & $40.9$ \\
ZwZ-4B\textsuperscript{\dag}~\citep{wei2026zwz} & $92.7$ & $81.8$ & $79.5$ & $68.5$ & $68.1$ & $55.7$ \\
P2R-4B\textsuperscript{\ddag}~\citep{li2026perceivetoreason} & $93.2$ & $81.9$ & $80.5$ & $66.3$ & $66.8$ & $52.5$ \\
Imagine-OPD-4B\textsuperscript{\dag}~\citep{cai2026thinking} & $88.0$ & \textbf{83.3} & $78.6$ & $68.1$ & $67.6$ & $43.2$ \\
AXPO-4B~\citep{kang2026axpo} & $88.9$ & $80.9$ & $78.3$ & -- & -- & -- \\
\MethodName & \textbf{94.6} & $81.5$ & \textbf{80.8} & \textbf{68.7} & \textbf{68.6} & \textbf{66.3} \\
\midrule
\multicolumn{7}{l}{\textit{Based on Qwen3.5-4B}} \\
\midrule
VAD-4B\textsuperscript{\dag}~\citep{zhang2026vad} & $92.2$ & $85.4$ & $83.4$ & \textbf{77.0} & $71.5$ & $60.6$ \\
Vision-OPD-4B\textsuperscript{\dag}~\citep{yuan2026visionopd} & $92.2$ & $84.5$ & $80.4$ & $74.9$ & $70.8$ & $59.8$ \\
EAD-4B\textsuperscript{\dag}~\citep{hu2026ead} & $91.6$ & $86.4$ & $82.9$ & $72.4$ & $71.1$ & $60.8$ \\
V-Zero-4B\textsuperscript{\dag}~\citep{sun2026vzero} & $89.0$ & \textbf{87.8} & $82.6$ & -- & -- & $57.8$ \\
VPS-4B~\citep{tao2026vps} & $86.9$ & $85.8$ & $82.8$ & -- & -- & $61.7$ \\
VISTA-4B~\citep{jiang2026vista} & $88.5$ & $83.1$ & $79.1$ & -- & -- & -- \\
\MethodName & \textbf{95.1} & $86.2$ & \textbf{84.5} & $73.7$ & \textbf{71.9} & \textbf{66.4} \\
\bottomrule
\end{tabular}
}
\caption{Answer accuracy (\%) of models based on Qwen3-VL-4B and Qwen3.5-4B. Results of Qwen3-VL-4B-Instruct are taken from \citet{wei2026zwz}, and the Qwen3.5-4B baselines use the scores reported in their papers. With Qwen3-VL-4B, \MethodName starts from Qwen3-VL-4B-Thinking. $^\dag$ denotes models that do not crop or zoom into the image at inference time. $^\ddag$P2R-4B does not call a tool during reasoning; it first predicts region boxes, and then answers from the image with crops of these boxes.}
\label{tab:qwen3vl-main}
\label{tab:qwen35-main}
\end{table*}

\section{More Results and Analysis}
\label{app:more-results}

\subsection{Selection Pass Rate During Training}
\label{app:selection-rate}

At each RL step, selection checks candidate questions from the synthesised dataset with the current policy and trains on those that pass (\cref{sec:selection}). \Cref{fig:selection-rate} reports the pass rate, i.e., the number of synthesised questions trained on at a step divided by the number checked at that step. For Qwen2.5-VL-7B, the pass rate stays between $17.9\%$ and $27.9\%$ over the 40 training steps, and $10{,}901$ of $48{,}299$ checked questions ($22.6\%$) pass. Selection checks somewhat more questions per step later in training to fill a batch, from $966$--$1{,}339$ in the first ten steps to $1{,}034$--$1{,}633$ in the last ten. For Qwen3-VL-4B (\cref{app:qwen3vl}), the pass rate stays between $18.9\%$ and $30.2\%$ during training, and $10{,}580$ of $42{,}948$ checked questions ($24.6\%$) pass.

\subsection{Ablation Analysis}
\label{app:ablation-analysis}

\Cref{tab:ablation-full} reports the ablation models on more benchmarks. On HR-Bench 4K, removing selection or synthesised data lowers accuracy from $79.5\%$ to $75.1\%$ and $75.6\%$, consistent with HR-Bench 8K, whereas removing unanswerable data leaves it unchanged ($79.5\%$). MME-RealWorld varies less across the ablations, from $62.9\%$ to $65.9\%$ on English and from $63.8\%$ to $67.1\%$ on Chinese.

\begin{table*}[t]
\centering
\resizebox{0.88\textwidth}{!}{
\begin{tabular}{lrrrrrrrrrr}
\toprule
 & \multicolumn{3}{c}{Controlled, human} & \multicolumn{3}{c}{Controlled, judge} & \multicolumn{4}{c}{Real crops, human} \\
\cmidrule(lr){2-4}\cmidrule(lr){5-7}\cmidrule(lr){8-11}
Model & FNR & FPR & Acc. & FNR & FPR & Acc. & Crop contains target & FNR & FPR & Acc. \\
\midrule
DeepEyes & $0.0$ & $100.0$ & $50.0$ & $0.0$ & $96.2$ & $51.9$ & $66.3$ & $0.0$ & $87.2$ & $80.8$ \\
PixelReasoner & $0.0$ & $80.8$ & $59.6$ & $0.0$ & $73.1$ & $63.5$ & $52.4$ & $7.3$ & $91.3$ & $54.2$ \\
CodeV & $84.5$ & $3.4$ & $56.0$ & $87.9$ & $0.0$ & $56.0$ & $67.5$ & $77.3$ & $12.7$ & $34.6$ \\
\MethodName & $12.1$ & $6.9$ & $90.5$ & $19.0$ & $0.0$ & $90.5$ & $90.9$ & $2.1$ & $12.1$ & $97.1$ \\
\bottomrule
\end{tabular}
}
\caption{Human evaluation of first-crop target recognition on V* (\%). Controlled: ground-truth crops versus random crops, grey images, and black images, labelled by humans and by the judge on the same samples. Real crops: the models' own first crops; ``Crop contains target'' is the percentage of crops that annotators judged to contain the target.}
\label{tab:human-eval}
\end{table*}

\begin{table*}[t]
\centering\small
\begin{tabular}{lrrrrrr}
\toprule
 & \multicolumn{3}{c}{Wrong region} & \multicolumn{3}{c}{Grey or black image} \\
\cmidrule(lr){2-4}\cmidrule(lr){5-7}
Model & Not noticed & Zoom again & Other & Not noticed & Zoom again & Other \\
\midrule
DeepEyes & $90.9$ & $0.0$ & $9.1$ & $100.0$ & $0.0$ & $0.0$ \\
PixelReasoner & $85.3$ & $5.9$ & $8.8$ & $100.0$ & $0.0$ & $0.0$ \\
CodeV & $21.2$ & $76.2$ & $2.6$ & $7.9$ & $92.1$ & $0.0$ \\
\MethodName & $7.6$ & $92.4$ & $0.0$ & $0.0$ & $100.0$ & $0.0$ \\
\bottomrule
\end{tabular}
\caption{Model behaviour when the first crop, or its replacement, does not contain the target, based on human annotation (\%). ``Wrong region'' refers to a first crop that misses the target, which is either the model's own wrong crop or a random crop that replaces it. ``Not noticed'' means that the model responds as if the target were in the image; ``Other'' includes answering from the original image, stating that the question is unanswerable, and guessing an answer after noticing that the target is missing.}
\label{tab:human-eval-behaviour}
\end{table*}

\subsection{Results with Qwen3-VL-4B and Qwen3.5-4B}
\label{app:qwen3vl}

\paragraph{Setup.}
To test whether our method transfers to other base models, we apply it to Qwen3-VL-4B-Thinking~\citep{bai2025qwen3vl} and Qwen3.5-4B~\citep{qwen2026qwen35}, with the same training data, data selection, and rewards as for Qwen2.5-VL-7B, and evaluate it in the same setting. For each base model, we compare against other models built on it, using their published scores where available.

\paragraph{Qwen3-VL-4B.}
We compare against Qwen3-VL-4B-Instruct, Qwen3-VL-4B-Thinking, and models built on Qwen3-VL-4B: ZwZ-4B~\citep{wei2026zwz}, P2R-4B~\citep{li2026perceivetoreason}, Imagine-OPD-4B~\citep{cai2026thinking}, and AXPO-4B~\citep{kang2026axpo}. We use published scores where available and evaluate the remaining cells with each model's own prompt, tool, and inference settings. AXPO-4B has no released weights, so only its published scores are listed. As shown in \cref{tab:qwen3vl-main}, \MethodName achieves the highest accuracy on five of the six benchmarks: V* ($94.6\%$), HR-Bench 8K ($80.8\%$), MME-RealWorld English ($68.7\%$) and Chinese ($68.6\%$), and ZoomBench ($66.3\%$), where it exceeds the strongest baseline, ZwZ-4B, by $10.6$ points. On HR-Bench 4K, it obtains $81.5\%$, lower than Imagine-OPD-4B ($83.3\%$), P2R-4B ($81.9\%$), and ZwZ-4B ($81.8\%$). The listed HR-Bench 4K score of Imagine-OPD-4B is its published value; in our evaluation with its released settings, it obtains $75.3\%$.

\paragraph{Qwen3.5-4B.}
We compare against models built on Qwen3.5-4B: VAD-4B~\citep{zhang2026vad}, Vision-OPD-4B~\citep{yuan2026visionopd}, EAD-4B~\citep{hu2026ead}, V-Zero-4B~\citep{sun2026vzero}, VPS-4B~\citep{tao2026vps}, and VISTA-4B~\citep{jiang2026vista}, using the scores reported in their papers. As shown in \cref{tab:qwen35-main}, \MethodName achieves the highest accuracy on four of the six benchmarks: V* ($95.1\%$), HR-Bench 8K ($84.5\%$), MME-RealWorld Chinese ($71.9\%$), and ZoomBench ($66.4\%$), where it exceeds the strongest baseline, VPS-4B, by $4.7$ points. On HR-Bench 4K, it obtains $86.2\%$, lower than V-Zero-4B ($87.8\%$) and EAD-4B ($86.4\%$). On MME-RealWorld English, it obtains $73.7\%$, lower than VAD-4B ($77.0\%$) and Vision-OPD-4B ($74.9\%$). Four of the baselines are trained by on-policy distillation and answer without tools.

\subsection{Human Evaluation}
\label{app:human-eval}

We conduct a human evaluation to verify the automatic judge used in \cref{sec:diagnosis} and to examine how the models behave when the first crop does not contain the target. Four annotators labelled the response right after the first zoom, where the returned image is the model's own crop, a ground-truth crop centred on the target, a random crop, a grey image, or a black image. Without knowing the model or the image type, they labelled whether the image shows the target, whether the model says that the target is in the image, and, if the image does not show the target, what the model does next. Each annotator labelled 192 samples, 32 of which were shared by all annotators, giving 672 distinct samples.

\paragraph{Agreement and judge consistency.}
On the shared samples, Fleiss' kappa is $0.89$ for whether the image shows the target and $0.79$ for whether the model claims the target. Annotators agree on every sample with a replaced image, as in the controlled setting used in \cref{sec:diagnosis}; all disagreements occur on the models' own crops, as in the real-crop setting (\cref{app:section3-protocol}), where the target may be only partly presented. On replaced images, the judge agrees with the human labels on $95.6\%$ of samples (Cohen's kappa $0.91$; \cref{tab:human-eval}), which supports the judge-based results in \cref{sec:diagnosis}. On the models' own crops, the judge is less reliable at deciding whether the crop shows the target (Cohen's kappa $0.44$), which is why the main paper uses the controlled setting.

\paragraph{Behaviour when the first crop misses the target.}
When the image shows a wrong region, \MethodName notices that the target is not in image and zooms again in $92.4\%$ of cases, and it does so for every grey or black image (\cref{tab:human-eval-behaviour}). DeepEyes and PixelReasoner instead continue as if the target were present in $85.3\%$--$100\%$ of these cases. CodeV also zooms again, but it does so even when the crop contains the target (FNR $84.5\%$), so its retries do not show that it has checked the image.

\paragraph{Stating that the question is unanswerable.}
\MethodName never states that the question is unanswerable directly after a crop that misses the target; it does so only after zooming again. On its own crops that miss the target, it zooms again and then states that the question is unanswerable in $48.5\%$ of cases, which may happen when its later zooms also fail to find the target. On replaced images, only the first crop is replaced, so its next zoom returns a real crop of the image, where it usually finds the target and answers. It therefore states that the question is unanswerable on only $4.2\%$ of random crops and never on grey or black images.

\subsection{Answers Inferred Before the First Crop}
\label{app:first-turn-guess}

\begin{table}[t]
\centering\small
\begin{tabular}{lcc}
\toprule
Model & \begin{tabular}{@{}c@{}}Infers an answer\\before crop\end{tabular} & \begin{tabular}{@{}c@{}}Final prediction\\matches\end{tabular} \\
\midrule
DeepEyes & $88.0$ & $95.2$ \\
PixelReasoner & $21.8$ & $96.8$ \\
CodeV & $8.2$ & $100.0$ \\
\MethodName & $20.4$ & $89.5$ \\
\bottomrule
\end{tabular}
\caption{How often a model decides its answer before seeing the returned crop on V*. \textit{Infers an answer before crop}: percentage of questions for which the model infers an answer in its first reasoning turn, before the tool call. \textit{Final prediction matches}: among these questions, percentage for which the final prediction is the same as this answer.}
\label{tab:first-turn-guess}
\end{table}

\Cref{sec:diagnosis} and \cref{app:human-eval} show that DeepEyes and PixelReasoner often respond as if the target were present, even when the returned crop does not contain it. Checking their outputs, we notice that models sometimes infer an answer in their first reasoning turn, before they call the tool. For example, for the question ``What is the color of the cart?'', DeepEyes first writes ``The cart in the image is green. I can see that the cart has a green body and green handles. Therefore, the correct answer is D. green.'' and then calls the tool to zoom into a region. Thus, it has chosen its answer before it sees the crop, which may be one reason for its unfaithful responses.

To measure how often this happens, we use the judge to determine whether the first reasoning turn $z_1$ states or clearly implies an answer. In a separate session, the judge checks whether this answer matches the model's final prediction.
As shown in \cref{tab:first-turn-guess}, DeepEyes infers an answer before the crop for $88.0\%$ of questions, and its final prediction matches this answer for $95.2\%$ of them. In other words, DeepEyes has already decided its answer before it sees the crop in $83.7\%$ of cases. This supports the possible reason above for DeepEyes and agrees with the low attention it gives to the first crop (\cref{app:attention}). These results do not explain the behaviour of PixelReasoner and CodeV, which infer an answer before the first crop for only $21.8\%$ and $8.2\%$ of questions, so their unfaithful responses could have other causes. \MethodName infers an answer before the crop for $20.4\%$ of questions.

\subsection{Attention on the First Crop}
\label{app:attention}

\begin{figure}[t]
\centering
\includegraphics[width=\linewidth]{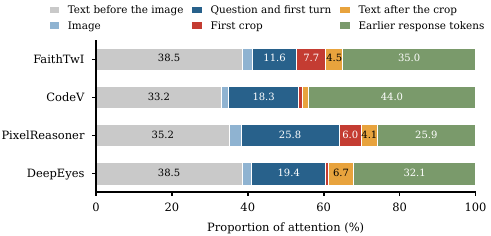}
\caption{Proportion of attention over the input while the model writes its response after the first crop (\%).}
\label{fig:attention}
\end{figure}

We analyse how each model distributes its attention over the context while it generates its response after the first crop. For each response token, we split its attention over all earlier tokens into six segments: the text before the image, the image, the question and the first assistant turn, the first crop, the text after the crop, and the earlier response tokens. We then average the proportions over all layers, attention heads, and response tokens, and then over the V* examples. As shown in \cref{fig:attention}, \MethodName gives the first crop $7.7\%$ of its attention, compared with $6.0\%$ for PixelReasoner, $1.1\%$ for CodeV, and $0.7\%$ for DeepEyes. It also attends less to the question and its first turn, with $11.6\%$ compared with $18.3\%$--$25.8\%$ for the baselines. These results suggest that \MethodName relies more on the returned crop when it reasons after a zoom, which is consistent with its more faithful target recognition.

\onecolumn
\section{Prompts Used in the Method}
\label{app:method-prompts}

We present the prompts used for object localisation, question generation, question verification, unanswerable questions, adaptive data selection, and tool use. Placeholders such as \texttt{\{question\}} are filled for each example.

\newcounter{methodprompt}
\DefineVerbatimEnvironment{MethodPrompt}{Verbatim}{fontsize=\small,breaklines=true,breakanywhere=true,breaksymbolleft={},frame=single,framesep=3mm}

\subsection{Object localisation and question generation}
\label{app:prompt-generation}

\paragraph{Prompt 1: Listing object types.}
\refstepcounter{methodprompt}\label{prompt:discovery}
For each image tile, the question generator lists the visible object types, and SAM3 then localises objects of these types (\cref{sec:data-synthesis}).
\begin{MethodPrompt}
List the visible object categories in the attached image, including small objects, signs, labels, and text blocks.

Use short, concrete category names suitable for text-prompted segmentation, such as "bottle" or "price tag." Do not describe positions, list individual instances, transcribe text, or infer objects that are not visible. Include each category once, even when several instances appear.

Return a JSON object only, without Markdown fences:
{"objects": ["bottle", "price tag", "text block"]}

If no categories can be identified, return {"objects": []}.
\end{MethodPrompt}

\paragraph{Prompt 2: Single-object questions.}
\refstepcounter{methodprompt}\label{prompt:single-object}
The question generator receives the object crop $I_{o_i}$, the ground-truth crop $v^+_{o_i}$, and the downscaled whole image $I$, and proposes questions targeting one ability with draft answers.
\begin{MethodPrompt}
Design visual search questions: questions that require locating a target in an image and inspecting its details.

The agent will receive an unannotated image and a question, and can use tools to inspect image details. Design questions that require close inspection of the selected region to answer.

Write up to {num_questions} distinct questions supported by the evidence. Prefer different facts rather than paraphrases. Do not force different ability categories or invent facts to reach the requested count. If none of the allowed abilities fits this region, return {"questions": []} instead of changing the target or inventing a question.

The attached images appear in this order:
- Image 1: a detailed crop of the selected evidence region, enlarged when small and padded with white. Use the visible scene content to determine the answers; padding is not part of the scene.
- Image 2: a larger context crop around the selected region. Use it to understand the surroundings.
- Image 3: the whole image, resized, with the selected region outlined in red. Use it to identify the target within the full scene.

Write clear, self-contained questions that identify the target in the full image without referring to crops or annotations. Do not hint at the answer. Provide a concise answer supported by visible details in the selected region.

Before finalizing each question, inspect the whole image for other objects that fit its wording. If they could give different answers, add a visible position or nearby landmark that distinguishes the selected target without revealing the answer. Multiple matching objects are acceptable only when the requested answer is visibly the same for all of them; similar appearance alone is not enough. Omit the question if you cannot resolve conflicting or uncertain answers. Use the red outline only to locate the region; determine the object's color and appearance from the unmarked detailed crop.

For each question, label the main ability needed to answer it using only the following list. Omit questions that require an ability outside this list. Positional wording may locate the target, but questions about relationships between two objects are generated separately.
- text_recognition: read visible text or symbols.
- color_recognition: identify a visible color.
- entity_recognition: identify an object or entity.
- counting: count all visible items in a clearly defined group or area.
- structure_recognition: identify shape, parts, arrangement, or a visible pattern.
- material_recognition: identify what an object or part is made of when the material is visually distinguishable. Use broad material names; do not guess an exact composition or force a material question.

Additional instructions:
Each question is later checked against a small vision-language model: it is discarded if that model can answer it from the question text alone (for example the usual colour of such an object, an object whose name follows from a described shape or use, an obvious position), or from the whole image without zooming. That model sees the whole image at up to about 8 megapixels (about 3,300 x 2,400 pixels), so anything legible at that size is too easy: on documents, infographics and charts ask about the smallest print (fine print, footnotes, small labels, tiny numbers in a table or figure), never headings or normal body text; in photos ask about small or distant things. Ask about a detail that has to be seen up close and differs from picture to picture: printed text or numbers, a colour, pattern or material that is not the typical one for such an object, how many small items there are, or a fine part or marking. Refer to the target by where it is and what is around it, never by features that give the answer away.

The image analysis model provided this region description for reference:
{region_description}

For every question, also write guess_without_image: the answer a person who cannot see the picture would most likely give from the question text and common sense. If that guess would be the same as the true answer, do not ask that question. Ask instead about a detail whose answer cannot be guessed: printed text or numbers, a colour, marking or pattern that is not the usual one for such an object, or how many small items there are.

These question-answer examples illustrate the allowed abilities using other images. Do not copy their answers or assume their objects, text, numbers, or relationships exist in the current image. Use them as inspiration, not templates to copy. You may use any listed ability, including those without an example. Base each question and answer on the current evidence. If no allowed ability is suitable, return an empty questions list.

{examples}

Return JSON containing at most {num_questions} question objects, without Markdown fences:
{"questions": [{"question": "...", "answer": "...", "ability": "one of the listed abilities", "guess_without_image": "the answer a person who cannot see the picture would most likely give"}]}

Return fewer questions when the evidence supports fewer, or {"questions": []} when none are suitable.
\end{MethodPrompt}

\paragraph{Prompt 3: Spatial-relation questions.}
\refstepcounter{methodprompt}\label{prompt:relation}
The question generator receives the object crop $I_{o_i}$, a larger crop around it with candidate objects marked by numbers, and the downscaled whole image $I$. It chooses a second object $o_j$ and proposes a question about the relation between the two objects.
\begin{MethodPrompt}
Design visual search questions about spatial relationships between two objects.

The agent will receive an unannotated image and a question, and can use tools to inspect image details. Design questions that require close inspection to identify the objects and determine their relationship.

Write up to {num_questions} distinct, open-ended questions, each about object A and a different numbered object. Ask about a clearly visible relationship, such as relative position, containment, contact, or overlap.

The attached images appear in this order:
- Image 1: a detailed crop of object A, enlarged when small and padded with white.
- Image 2: a larger context crop around A, with A marked "A" and other objects marked with numbers. Use these numbers to select a second object.
- Image 3: the whole image, resized and unannotated. Use it to identify both objects within the full scene.

Write clear, self-contained questions that identify both objects in the full image without referring to crops or annotations. Do not hint at the answer or offer answer choices. Provide a concise answer supported by the visible relationship between the objects.

Before finalizing each question, inspect the whole image for other objects that could fit either description. If different matching pairs could give different answers, add visible positions or landmarks that distinguish the intended pair without stating the relationship being asked about. Multiple matching pairs are acceptable only when the answer is visibly the same for every pair. Omit the question if you cannot resolve conflicting or uncertain answers. Use the added marks only to select objects; determine their appearance from the original scene content.

These question-answer examples illustrate the allowed abilities using other images. Do not copy their answers or assume their objects, text, numbers, or relationships exist in the current image. Use them as inspiration, not templates to copy. You may use any listed ability, including those without an example. Base each question and answer on the current evidence. If no allowed ability is suitable, return an empty questions list.

{examples}

For each question, return the selected object's number as second_object, a brief name for A as a, and a brief name for the selected object as b. Name objects by their identity or appearance without revealing their spatial relationship.

Return JSON only, without Markdown fences:
{"questions": [{"second_object": "1", "a": "striped cup", "b": "silver kettle", "question": "Where is the striped cup relative to the silver kettle?", "answer": "To the left of the kettle."}]}

Return fewer questions when fewer pairs support a clear question and answer, or {"questions": []} when none are suitable.

Describe each of the two objects in the question so that exactly one object in the whole image fits the description: add its colour, printed text, size, or position relative to a large and unique landmark. A bare name such as "the boat" or "the headlight" is not enough when the image contains several of them. Give a and b as these full descriptions, exactly as they appear in the question (for example "the red rowing boat beside the pier", not "boat").

The image analysis model provided this description of region A for reference:
{region_description}
\end{MethodPrompt}

\subsection{Question verification}
\label{app:prompt-answers}

\paragraph{Prompt 4: Target identification.}
\refstepcounter{methodprompt}\label{prompt:target-match}
Numbered boxes mark the objects localised by SAM3 in the region. The question generator selects every object that matches the target described in the question. For single-object questions, \texttt{\{description\}} is ``the object or area asked about''; for spatial-relation questions, it is the description of the object being checked.
\begin{MethodPrompt}
Image 1 shows the whole scene. Image 2 shows a local view with numbered objects; white numbers on blue tags connect to boxes by lines. Identify every numbered object matching the target in the complete question. Interpret positions in the whole image, not relative to the local crop. The target description only identifies which role to check; all locators in the question still apply. Do not answer the question or infer the target from an assumed answer. For a group or counting question, a box must contain the entire specified group or area, not just one member. Return no match if no box covers that scope, or if the candidate is only a clipped fragment that cannot be identified.

Target: {description}
Question: {question}

Return only JSON with a matches list of integers, for example {"matches": [1, 3]} or {"matches": []}. Do not include Markdown or explanatory text.
\end{MethodPrompt}

\paragraph{Prompt 5: Clarifying the target description (single-object questions).}
\refstepcounter{methodprompt}\label{prompt:target-rewrite}
If the question generator does not select exactly the intended target, it revises the question given the intended object and the matching result.
\begin{MethodPrompt}
Rewrite this question with the smallest added position or neighbouring landmark that identifies the same target in the unmarked whole scene. Image 1 is the whole image; the remaining images show the numbered local views for the failed references. Object numbers belong only to their specified image. Preserve the original target, asked property, and answer; do not redirect the question to a neighbour or change a group question into a question about one member. If the target does not support the original question, leave it unchanged. Do not add answer clues or refer to helper boxes, numbers, crops, or marks. Return only JSON: {"question": "..."}.

Use this feedback to locate the reference problem, but confirm it against the images because the previous checker may be mistaken. Distinguish the intended target from other plausible matches when needed; do not merely describe the object the checker selected.

{failure_feedback}

Question: {question}
\end{MethodPrompt}

\paragraph{Prompt 6: Clarifying the target descriptions (spatial-relation questions).}
\refstepcounter{methodprompt}\label{prompt:target-rewrite-relation}
For spatial-relation questions, the prompt also gives the current descriptions of the two objects, and the response contains their revised descriptions.
\begin{MethodPrompt}
Rewrite this question with the smallest added position or neighbouring landmark that identifies the same target in the unmarked whole scene. Image 1 is the whole image; the remaining images show the numbered local views for the failed references. Object numbers belong only to their specified image. Preserve the original target, asked property, and answer; do not redirect the question to a neighbour or change a group question into a question about one member. If the target does not support the original question, leave it unchanged. Do not add answer clues or refer to helper boxes, numbers, crops, or marks. Current roles: A = {description_a}; B = {description_b}. Preserve which physical object occupies each role and update both descriptions to agree with the revised question. Return only JSON: {"question": "...", "objects": {"a": "updated description of A", "b": "updated description of B"}}.

Use this feedback to locate the reference problem, but confirm it against the images because the previous checker may be mistaken. Distinguish the intended target from other plausible matches when needed; do not merely describe the object the checker selected.

{failure_feedback}

Question: {question}
\end{MethodPrompt}

\paragraph{Prompt 7: Answering from the image.}
\refstepcounter{methodprompt}\label{prompt:reanswer}
We sample multiple answers using the whole image $I$, the object crops, and the ground-truth crops $V^+$, without the draft answer.
\begin{MethodPrompt}
Answer the question using the attached images: detailed crops, a larger context view, and the whole image. Use the context and whole image to identify the objects, and the detailed crops to inspect them. Outlines on the whole image are added location markers, not part of the scene; determine colors and appearance from the unmarked crops. Reply with a concise answer only.

Question: {question}
\end{MethodPrompt}

\paragraph{Prompt 8: Final reference answer.}
\refstepcounter{methodprompt}\label{prompt:relabel}
The question generator reviews the sampled answers alongside the draft answer and proposes the final reference answer $y$. An empty answer means that the visual evidence is insufficient, and the question is discarded.
\begin{MethodPrompt}
Answer the question using the attached detailed crops, context view, and whole image. Review the proposed answer and independently collected answers below, then choose or combine their supported details into one concise answer. Use the images to resolve disagreements; repetition does not make an answer correct. Do not combine contradictory claims or add unsupported details. Outlines are location markers, not part of the scene; read colors and appearance from the unmarked crops. If the evidence is insufficient to resolve the answer, return an empty answer string.

Question: {question}
Proposed answer: {answer}
Collected answers: {samples}

Return only a JSON object with one string field: {"answer": "..."}.
\end{MethodPrompt}

\paragraph{Prompt 9: Answering for other matching objects.}
\refstepcounter{methodprompt}\label{prompt:reanswer-alt}
When target identification finds other objects that match the question, the two answer checks above are repeated on crops of these objects, with an added instruction to answer about them. The question is kept only if these answers agree with the reference answer.
\begin{MethodPrompt}
This check concerns alternative matching targets. Apply the question to the objects in the attached detailed crops, even if a locator in the question would lead you to a different object in the overview. Answer about these cropped objects, not the original target elsewhere in the scene. For a two-object relation, the first crop supplies the first object and the second crop supplies the second object; use their outlined positions in the overview to determine the relationship. If the requested detail cannot be established for these objects, say that it cannot be determined.

Answer the question using the attached images: detailed crops, a larger context view, and the whole image. Use the context and whole image to identify the objects, and the detailed crops to inspect them. Outlines on the whole image are added location markers, not part of the scene; determine colors and appearance from the unmarked crops. Reply with a concise answer only.

Question: {question}
\end{MethodPrompt}

\paragraph{Prompt 10: Reference answer for other matching objects.}
\refstepcounter{methodprompt}\label{prompt:relabel-alt}
In this check, \texttt{\{answer\}} is replaced by ``Not supplied for the alternative targets.''
\begin{MethodPrompt}
This check concerns alternative matching targets. Apply the question to the objects in the attached detailed crops, even if a locator in the question would lead you to a different object in the overview. Answer about these cropped objects, not the original target elsewhere in the scene. For a two-object relation, the first crop supplies the first object and the second crop supplies the second object; use their outlined positions in the overview to determine the relationship. If the requested detail cannot be established for these objects, say that it cannot be determined.

Answer the question using the attached detailed crops, context view, and whole image. Review the proposed answer and independently collected answers below, then choose or combine their supported details into one concise answer. Use the images to resolve disagreements; repetition does not make an answer correct. Do not combine contradictory claims or add unsupported details. Outlines are location markers, not part of the scene; read colors and appearance from the unmarked crops. If the evidence is insufficient to resolve the answer, return an empty answer string.

Question: {question}
Proposed answer: {answer}
Collected answers: {samples}

Return only a JSON object with one string field: {"answer": "..."}.
\end{MethodPrompt}

\paragraph{Prompt 11: Guessing without the image.}
\refstepcounter{methodprompt}\label{prompt:blind-generator}
The question generator answers each question four times without the image, and we reject questions that it answers correctly too often.
\begin{MethodPrompt}
You cannot see the picture this question is about. Give your single best guess anyway, using common sense and what such scenes usually look like. Reply with a short answer only.

Question: {question}
\end{MethodPrompt}

\subsection{Unanswerable questions}
\label{app:prompt-unanswerable}

\paragraph{Prompt 12: Answerability after editing.}
\refstepcounter{methodprompt}\label{prompt:edited-answerability}
After the evidence region is obscured, the question must no longer be answerable; after an edit away from the evidence, the original answer must be preserved.
\begin{MethodPrompt}
Image 1 shows a whole picture. The other images are detail views of parts of the same picture.

Question: {question}

Decide from these images alone whether the question can be answered. If they show what the question asks about clearly enough to give one definite answer, the verdict is "answerable" and you give that answer. If they do not (what the question asks about cannot be found, or cannot be made out well enough to answer with confidence), the verdict is "cannot be determined". Do not guess.

Answer with JSON only: {"verdict": "answerable" or "cannot be determined", "answer": "<short answer; empty when the verdict is cannot be determined>", "reason": "<one sentence>"}
\end{MethodPrompt}

\subsection{Adaptive data selection}
\label{app:prompt-selection}
To estimate the two accuracies in \cref{eq:selection}, the current policy answers each question with the whole image and with the whole image and the ground-truth crops. Both conditions use the system instruction ``You are a helpful assistant.'' The policy receives the question without its reference answer.

\paragraph{Prompt 13: Whole-image input.}
\refstepcounter{methodprompt}\label{prompt:selection-image}

\begin{MethodPrompt}
Answer the question using the attached image.

Question: {question}
Think first, then answer. Format strictly as:  <think>...</think>  <answer>...</answer>
\end{MethodPrompt}

\paragraph{Prompt 14: Whole image and ground-truth crop.}
\refstepcounter{methodprompt}\label{prompt:selection-crop}

\begin{MethodPrompt}
Answer the question using the two attached images.

Image 1 is the whole image. Image 2 is a crop of the relevant area, including its surroundings.
Use the whole image for context and the crop to inspect details.

Question: {question}
Think first, then answer. Format strictly as:  <think>...</think>  <answer>...</answer>
\end{MethodPrompt}

\subsection{Tool-use instruction}
\label{app:prompt-tool}

\paragraph{Prompt 15: System instruction.}
\refstepcounter{methodprompt}\label{prompt:tool-system}
This system instruction describes the zoom tool to the policy and how to use it.
\begin{MethodPrompt}
You are a helpful assistant.

# Tools
You may call one or more functions to assist with the user query.
You are provided with function signatures within <tools></tools> XML tags:
<tools>
{"type":"function","function":{"name":"image_zoom_in_tool","description":"Zoom in on a specific region of an image by cropping it based on a bounding box (bbox) and an optional object label.","parameters":{"type":"object","properties":{"bbox_2d":{"type":"array","items":{"type":"number"},"minItems":4,"maxItems":4,"description":"The bounding box of the region to zoom in, as [x1, y1, x2, y2], where (x1, y1) is the top-left corner and (x2, y2) is the bottom-right corner."},"label":{"type":"string","description":"The name or label of the object in the specified bounding box (optional)."}},"required":["bbox_2d"]}}}
</tools>

# How to call a tool
Return a json object with function name and arguments within <tool_call></tool_call> XML tags:
<tool_call>
{"name": <function-name>, "arguments": <args-json-object>}
</tool_call>

**Example**:
<tool_call>
{"name": "image_zoom_in_tool", "arguments": {"bbox_2d": [10, 20, 100, 200], "label": "the apple on the desk"}}
</tool_call>

# How to answer
First, reason inside <think></think>, following these steps:
1. Identify what detail the question asks for, and where in the original image it appears.
2. Check whether that detail is already clear in the original image. Zoom in on that region only if the detail is too small or unclear to answer from the original image.
3. If you zoomed, examine the zoomed image and check whether it contains the detail you were looking for. If it does, answer from it. If it does not, consider why: if you zoomed into the wrong region and the original image shows the object elsewhere, zoom in on that place once more; if the object or the detail is not in the original image, follow step 4.
4. Only when the original image does not contain what the question asks about, answer exactly: cannot be determined (or, if the question lists options, choose the option saying the image does not feature it). If the detail is there but small, look for it, zooming in if needed, and give your best reading; never invent a detail that is not there.

Then give the final answer inside <answer></answer>.
\end{MethodPrompt}

\paragraph{Prompt 16: User turn.}
\refstepcounter{methodprompt}\label{prompt:tool-user}
The first user turn contains the whole image, the question, and the instruction below.
\begin{MethodPrompt}
{image}
{question}
Think first following the steps above, call image_zoom_in_tool only if needed, then answer. Format strictly as:  <think>...</think>  <tool_call>...</tool_call> (if tools needed)  <answer>...</answer>
\end{MethodPrompt}

\paragraph{Prompt 17: Zoomed image returned by the tool.}
\refstepcounter{methodprompt}\label{prompt:tool-reply}
After each tool call, the zoomed image is returned in a new user turn; \texttt{\{zoomed image\}} marks its position, and \texttt{\{label\}} is the value of the optional \texttt{label} argument of the tool call (Prompt 15). When the call omits this argument, the text reads ``This is the region you zoomed into. Please examine it.''
\begin{MethodPrompt}
<tool_response>{zoomed image}
This is the region you zoomed into: "{label}". Please examine it.</tool_response>
\end{MethodPrompt}

\clearpage

\section{Data Synthesis Examples}
\label{app:curation-cases}
We illustrate the automated data synthesis pipeline with two single-object samples, one spatial-relation sample, and one unanswerable sample (\crefrange{fig:curation-colour}{fig:curation-removal}). The steps below show question generation, text-only guessing, target identification, and answer verification; the unanswerable sample also includes image editing and verification.

\paragraph{Single-object sample (\cref{fig:curation-colour}).}
\emph{Object sampling.} The sampled target is the waste bin beside the man in a white shirt and dark vest.

\emph{Question generation.} Using the object crop and scene context, the generator asks: ``At the outdoor market, the waste bin standing just to the left of the bald man in the white shirt and dark vest has a lower body of what color?'' It proposes ``Blue'' as the answer.

\emph{Text-only guessing.} None of the sampled answers without the image matches ``Blue'', so the question passes this check.

\emph{Target identification.} The generator selects object 8 in the Set-of-Mark image, confirming that the description identifies the intended bin.

\emph{Answer verification.} The generator reviews the proposed answer and the answers sampled from the image views, and returns ``Blue'' as the reference answer. The pipeline retains the question and answer.

\paragraph{Spatial-relation sample (\cref{fig:curation-spatial}).}
\emph{Object sampling.} The sampled target is the blue pedestrian-crossing sign.

\emph{Question generation.} The generator chooses the red circular sign as the second object from a numbered image. It asks: ``Where is the blue square pedestrian-crossing sign on the pole by the zebra crossing located relative to the small red circular sign on the blue building's facade?'' The proposed answer is ``Below it (and slightly to the right), near road level''.

\emph{Text-only guessing.} The question passes the guessing check, although some answers give the partial relation ``below'' without seeing the image.

\emph{Target identification.} The generator checks the two descriptions separately. It selects the blue sign as object 3 and the red sign as object 8 in their respective Set-of-Mark image. The red sign was object 6 in the partner-selection image; numbers are assigned separately in each image.

\emph{Answer verification.} The visual answers disagree on the small horizontal offset. After reviewing the images, the generator retains ``below and slightly to the right''. The consistency check also accepts partial answers stating only ``below'', so acceptance does not establish that every part of the answer requires visual inspection.

\paragraph{Single-object sample (\cref{fig:curation-entity}).}
\emph{Object sampling.} The sampled target is the hat worn by the pedestrian on the right of the square.

\emph{Question generation.} The generator asks: ``The pedestrian walking on the right side of the cobblestone square wears a light-colored hat; what style of hat is it, based on the shape of its brim?'' It proposes a bucket hat with a soft brim that slopes down all around. The scene identifies the pedestrian, while the close-up shows the brim.

\emph{Text-only guessing.} None of the sampled answers without the image matches the bucket-hat answer.

\emph{Target identification.} The generator selects object 10 in the Set-of-Mark image, confirming that the description identifies the intended hat.

\emph{Answer verification.} The sampled visual answers consistently identify a bucket hat, which the generator retains as the reference answer. The pipeline retains the question and answer.

\paragraph{Unanswerable sample (\cref{fig:curation-removal}).}
\emph{Object sampling.} The sampled target is the person in the courtyard behind the terracotta roof.

\emph{Question generation.} The generator asks: ``In the lower-left area of the hillside town, a person stands in a walled open space just behind a terracotta-tiled roof. Looking closely at the shirt this person is wearing, how does the color of the main torso panel compare with the color of the sleeves?'' The proposed answer describes a dark torso panel with lighter sleeves.

\emph{Text-only guessing.} None of the sampled answers without the image matches the proposed colour contrast.

\emph{Target identification.} The generator selects object 15 in the Set-of-Mark image, confirming the intended person.

\emph{Answer verification.} The visual answers support a darker torso and lighter sleeves, and the question passes the consistency check.

\emph{Editing and verification.} The pipeline covers the target region with a patch copied from elsewhere in the image. The generator judges that the edited image no longer supports the colour comparison, and the example is retained as unanswerable. This edit has a limitation: it removes the person and leaves visible seams, while the question still describes a person standing in the courtyard.

\begin{figure}[p]
\centering
\begin{subfigure}{\textwidth}
\centering
\includegraphics[width=\textwidth]{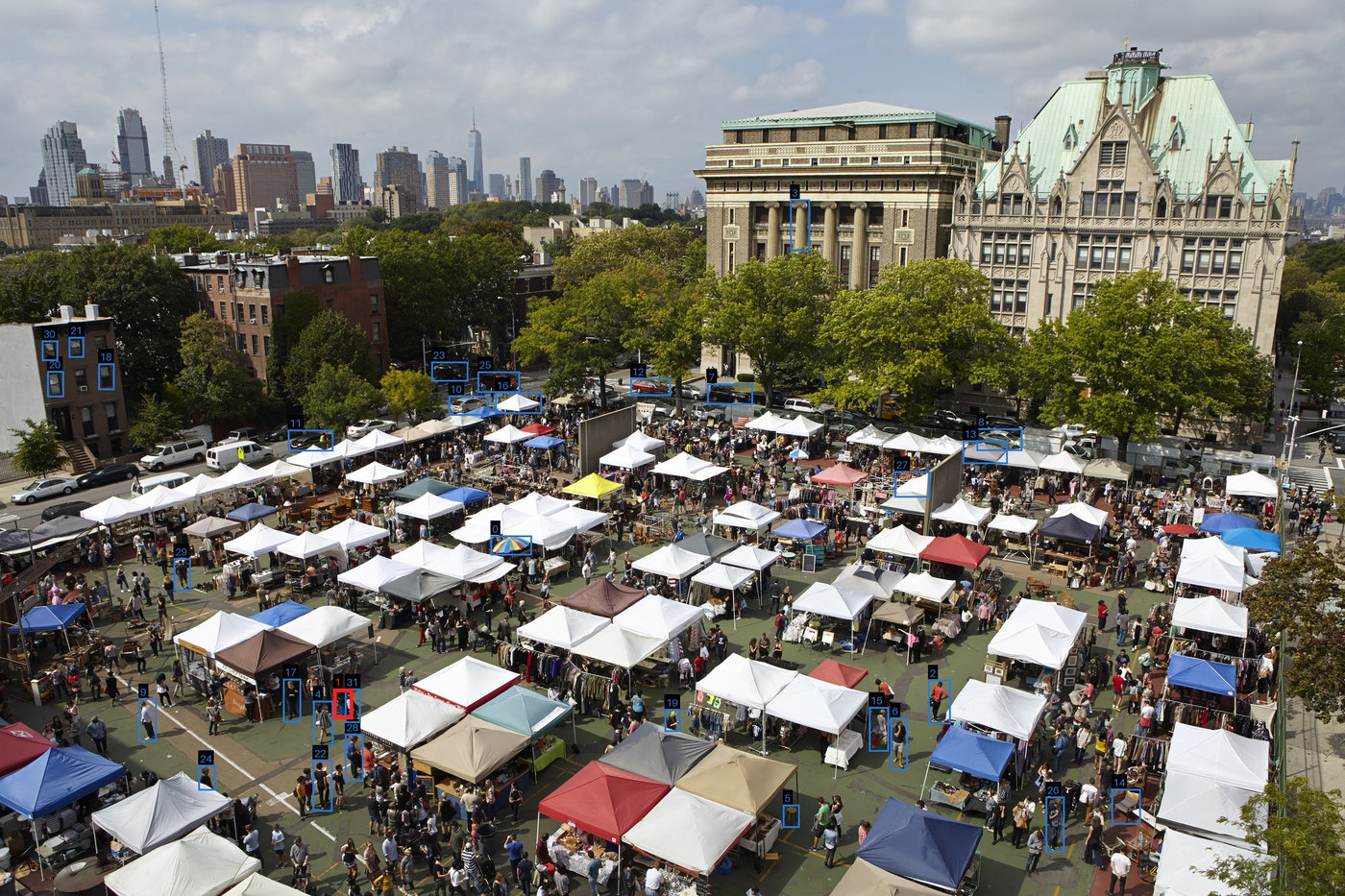}
\caption{Candidate objects detected by SAM~3. The red bounding box marks the sampled object.}
\label{fig:curation-colour-objects}
\end{subfigure}
\par\vspace{5pt}
\begin{subfigure}{\textwidth}
\centering
\includegraphics[width=.22\textwidth]{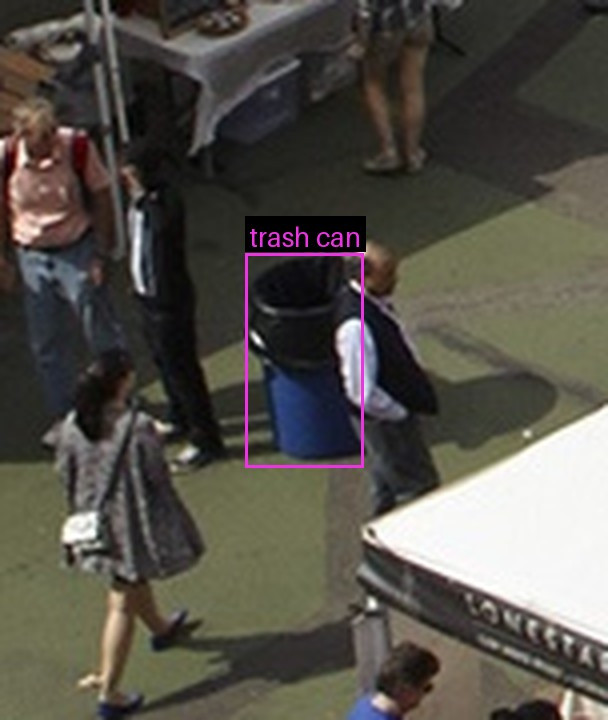}
\caption{The evidence crop shows the bin and its surroundings.}
\label{fig:curation-colour-evidence}
\end{subfigure}
\caption{Single-object sample: the sampled object and its colour.}
\label{fig:curation-colour}
\end{figure}

\begin{figure}[p]
\ContinuedFloat
\centering
\begin{subfigure}{\textwidth}
\centering
\includegraphics[width=\textwidth]{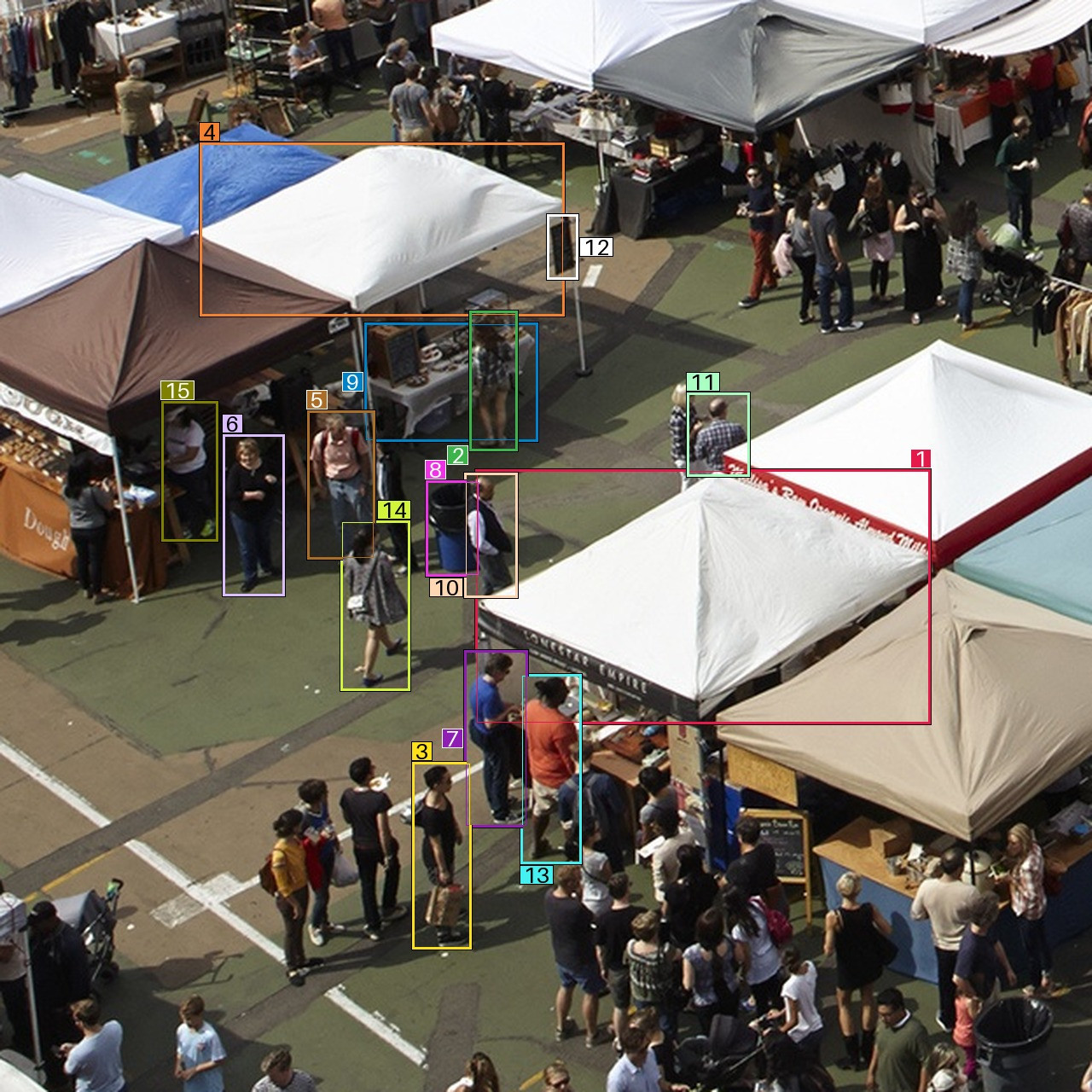}
\caption{The generator selects object 8, the waste bin described by the question.}
\label{fig:curation-colour-som}
\end{subfigure}
\caption{Single-object sample (continued): target identification using Set-of-Mark prompting.}
\end{figure}

\begin{figure}[p]
\centering
\begin{subfigure}{\textwidth}
\centering
\includegraphics[width=\textwidth]{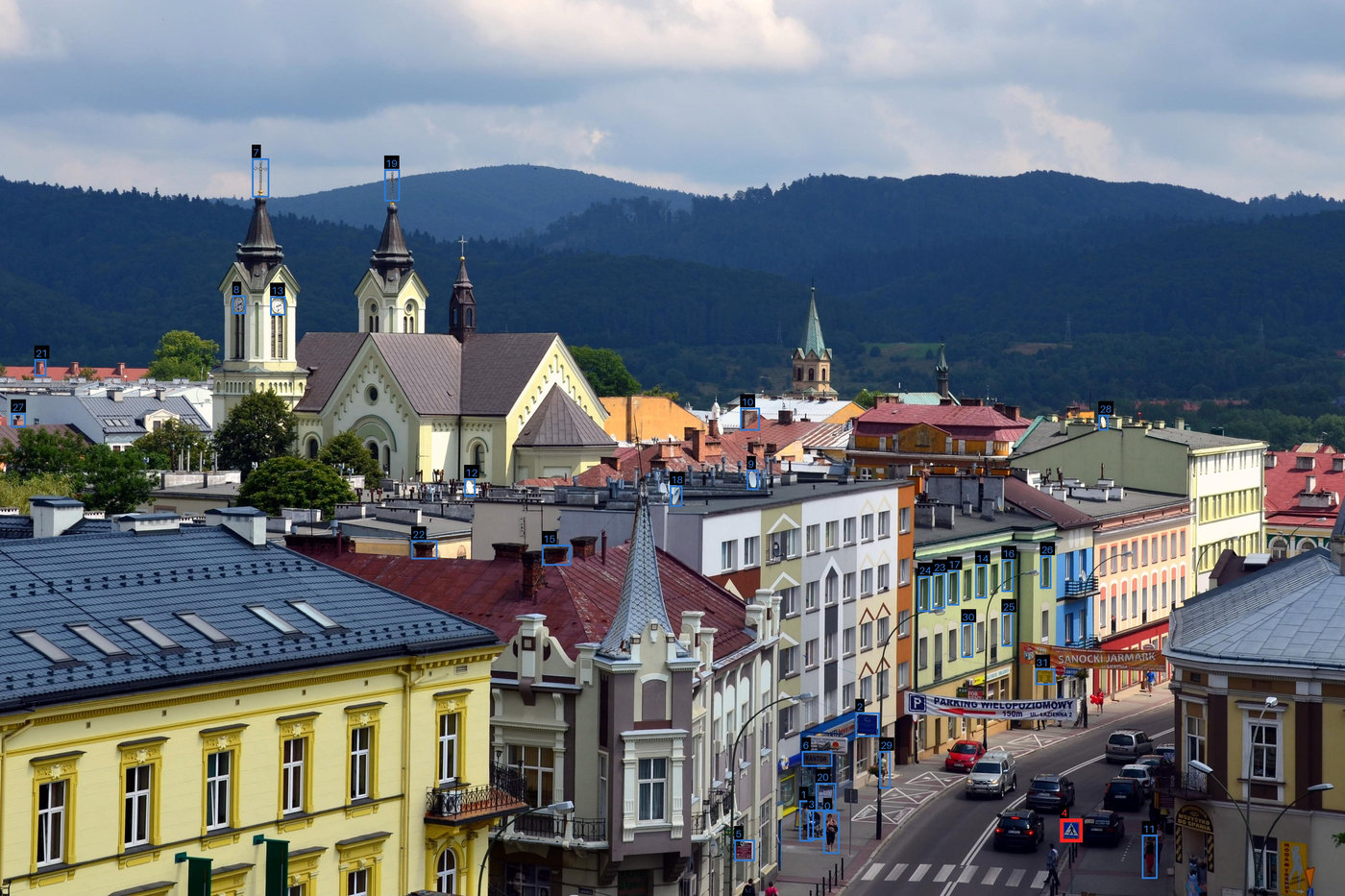}
\caption{Candidate objects detected by SAM~3. The red bounding box marks the sampled object.}
\label{fig:curation-spatial-objects}
\end{subfigure}
\par\vspace{5pt}
\begin{subfigure}{\textwidth}
\centering
\includegraphics[width=.18\textwidth]{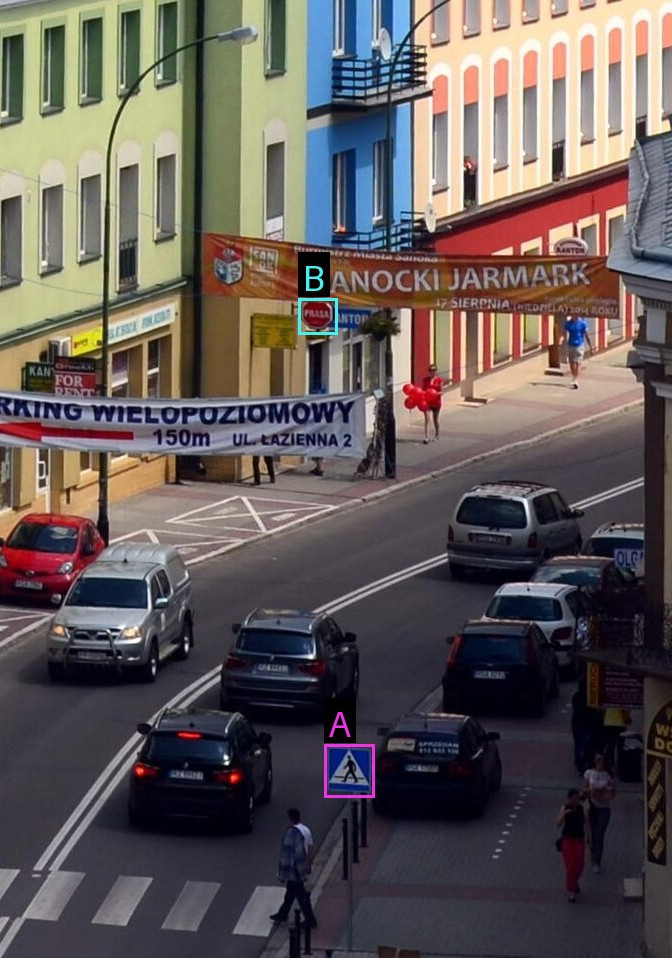}
\caption{The evidence crop contains both signs.}
\label{fig:curation-spatial-evidence}
\end{subfigure}
\caption{Spatial-relation sample: the sampled first object and the two target signs.}
\label{fig:curation-spatial}
\end{figure}

\begin{figure}[p]
\ContinuedFloat
\centering
\begin{subfigure}{\textwidth}
\centering
\includegraphics[width=\textwidth]{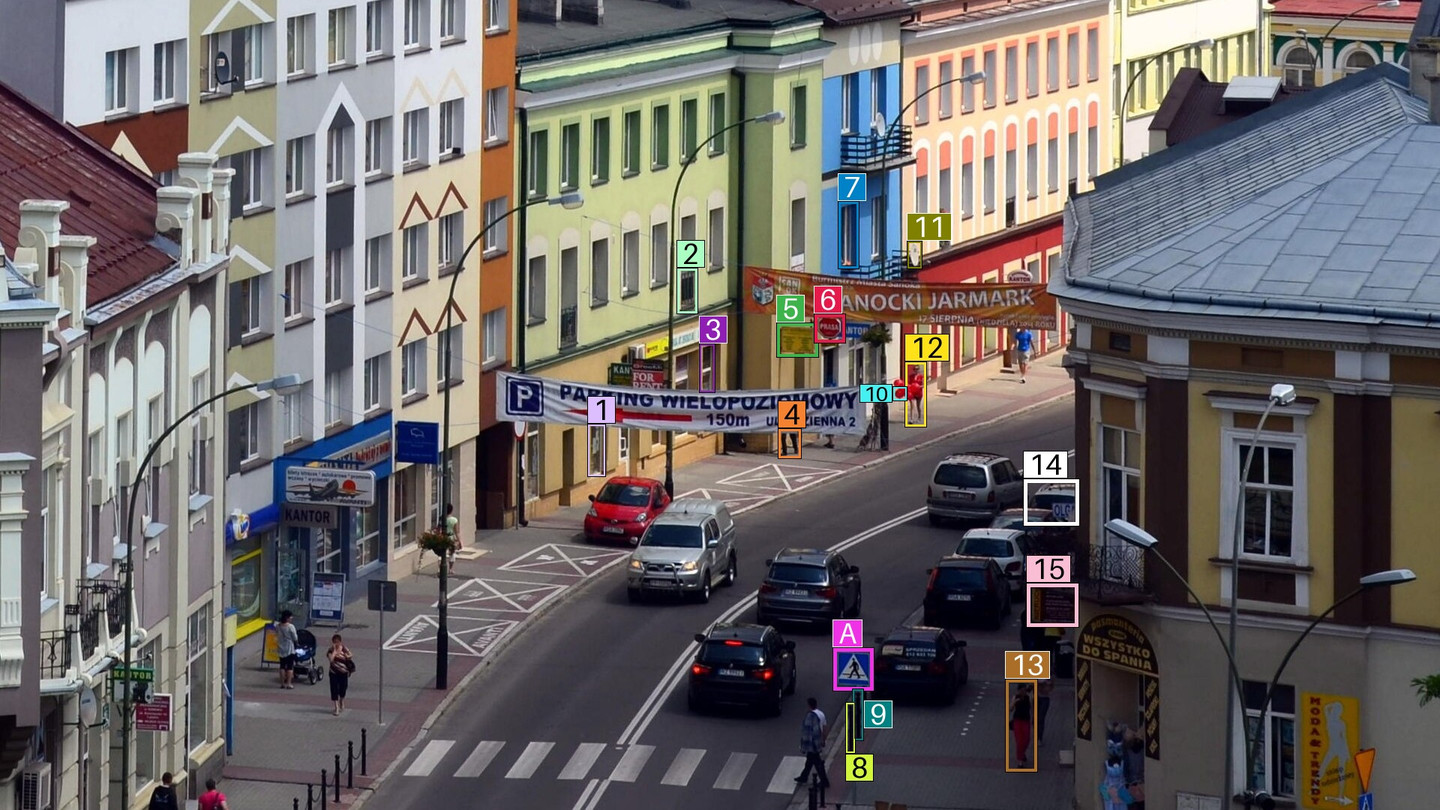}
\caption{The generator selects object 6, the red circular sign, as the second object.}
\label{fig:curation-spatial-partner}
\end{subfigure}
\caption{Spatial-relation sample (continued): choosing the second object.}
\end{figure}

\begin{figure}[p]
\ContinuedFloat
\centering
\begin{subfigure}{\textwidth}
\centering
\includegraphics[width=\textwidth]{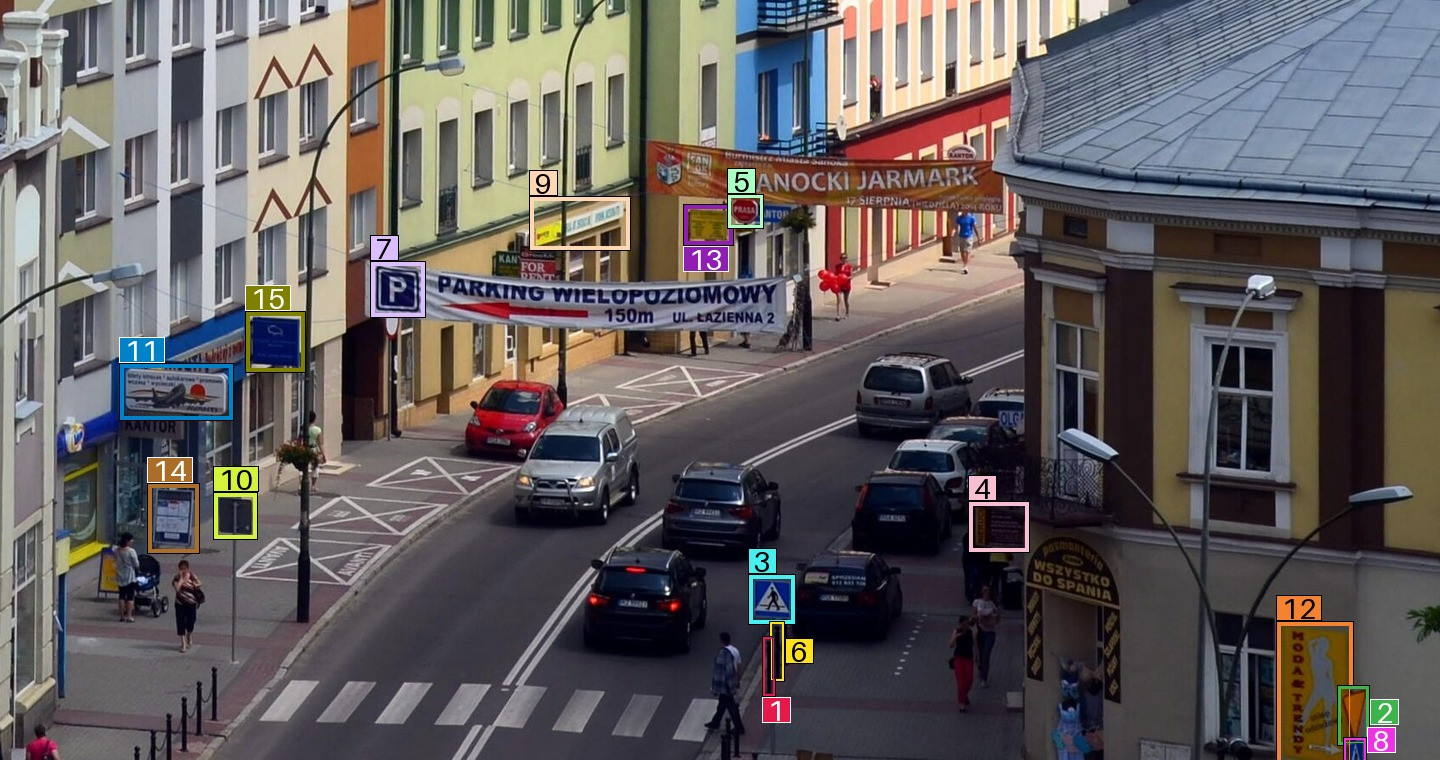}
\caption{The generator selects object 3, the blue pedestrian-crossing sign.}
\label{fig:curation-spatial-som-a}
\end{subfigure}
\par\vspace{5pt}
\begin{subfigure}{\textwidth}
\centering
\includegraphics[width=\textwidth]{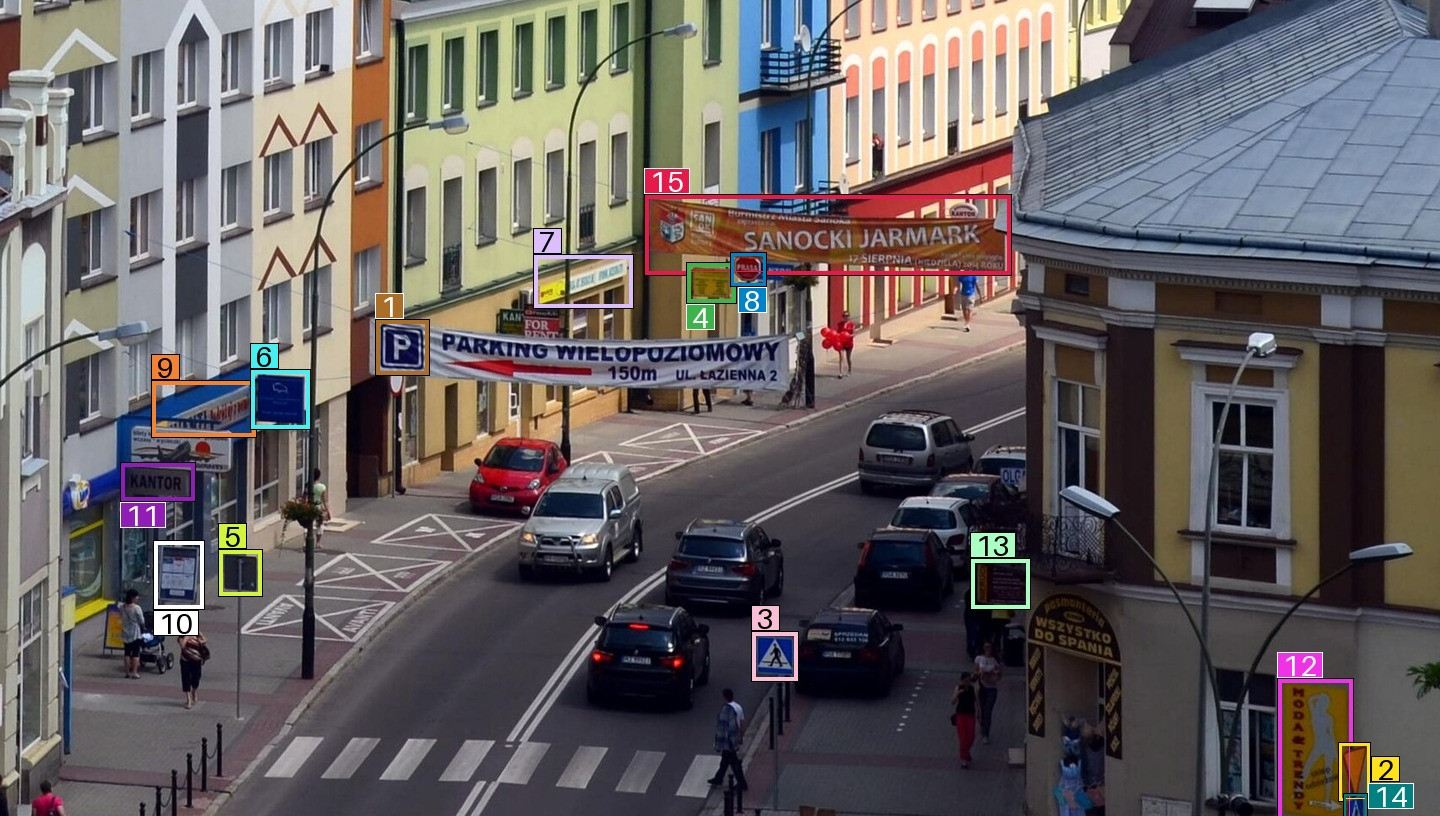}
\caption{The generator selects object 8, the red circular sign.}
\label{fig:curation-spatial-som-b}
\end{subfigure}
\caption{Spatial-relation sample (continued): checking each target description in a separate Set-of-Mark image.}
\end{figure}

\begin{figure}[p]
\centering
\begin{subfigure}{\textwidth}
\centering
\includegraphics[width=\textwidth]{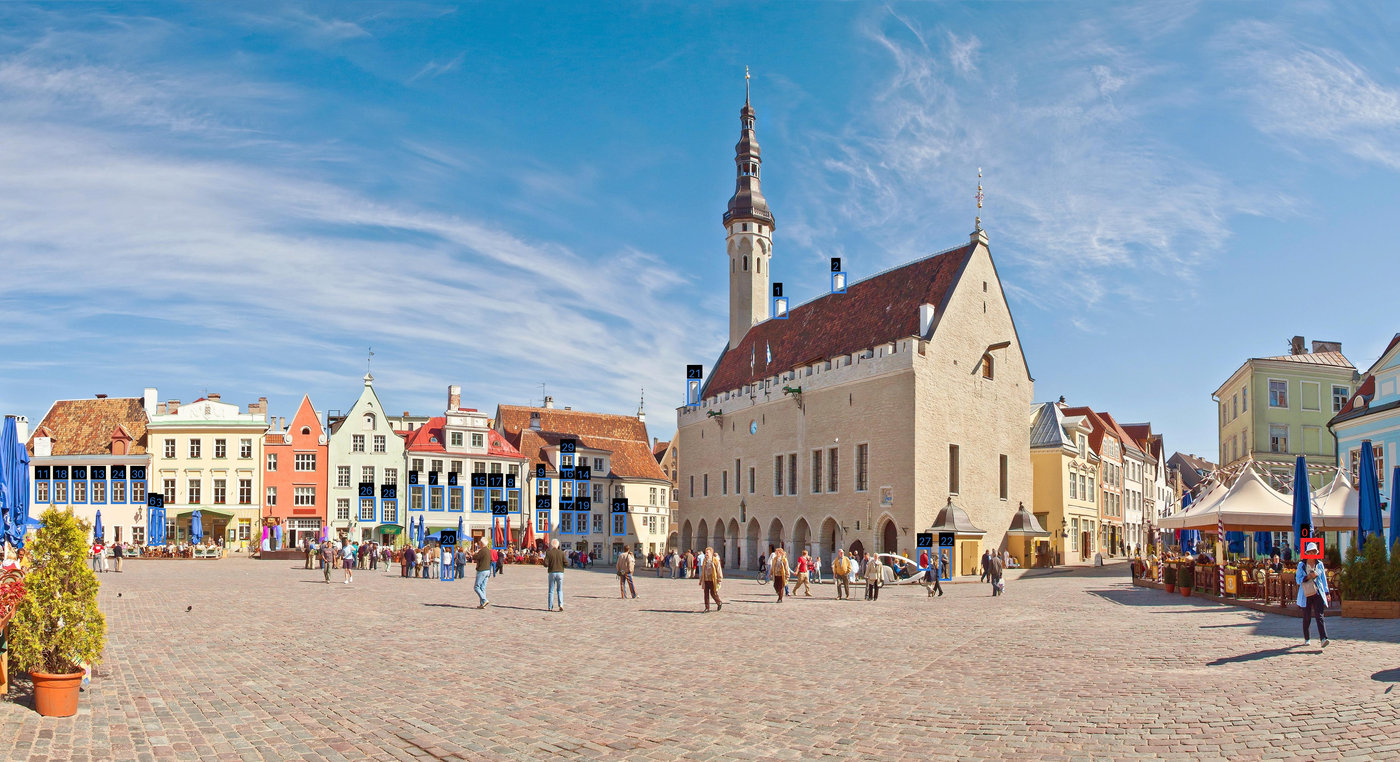}
\caption{Candidate objects detected by SAM~3. The red bounding box marks the sampled object.}
\label{fig:curation-entity-objects}
\end{subfigure}
\par\vspace{5pt}
\begin{subfigure}{\textwidth}
\centering
\includegraphics[width=.29\textwidth]{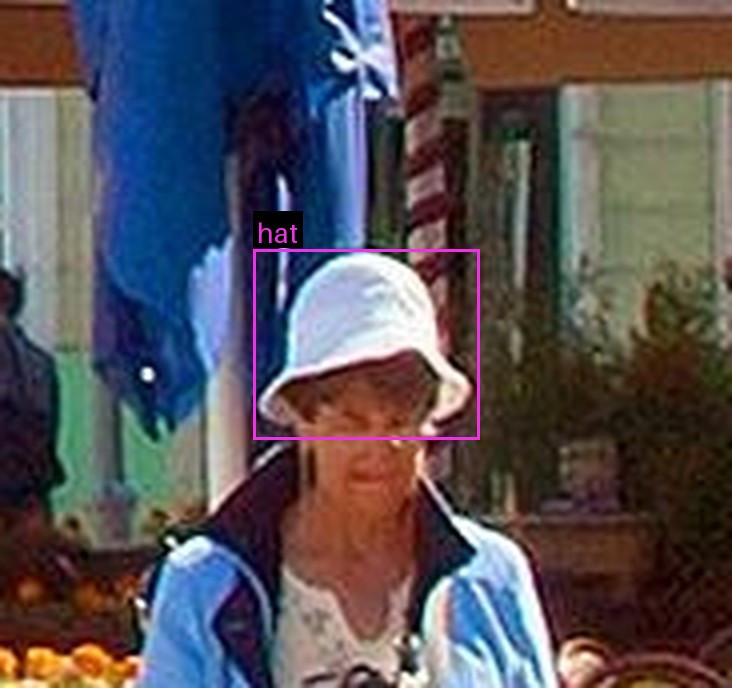}
\caption{The close-up shows the hat's downward-sloping brim.}
\label{fig:curation-entity-evidence}
\end{subfigure}
\caption{Single-object sample: the sampled hat and its shape.}
\label{fig:curation-entity}
\end{figure}

\begin{figure}[p]
\ContinuedFloat
\centering
\begin{subfigure}{\textwidth}
\centering
\includegraphics[width=\textwidth,height=.85\textheight,keepaspectratio]{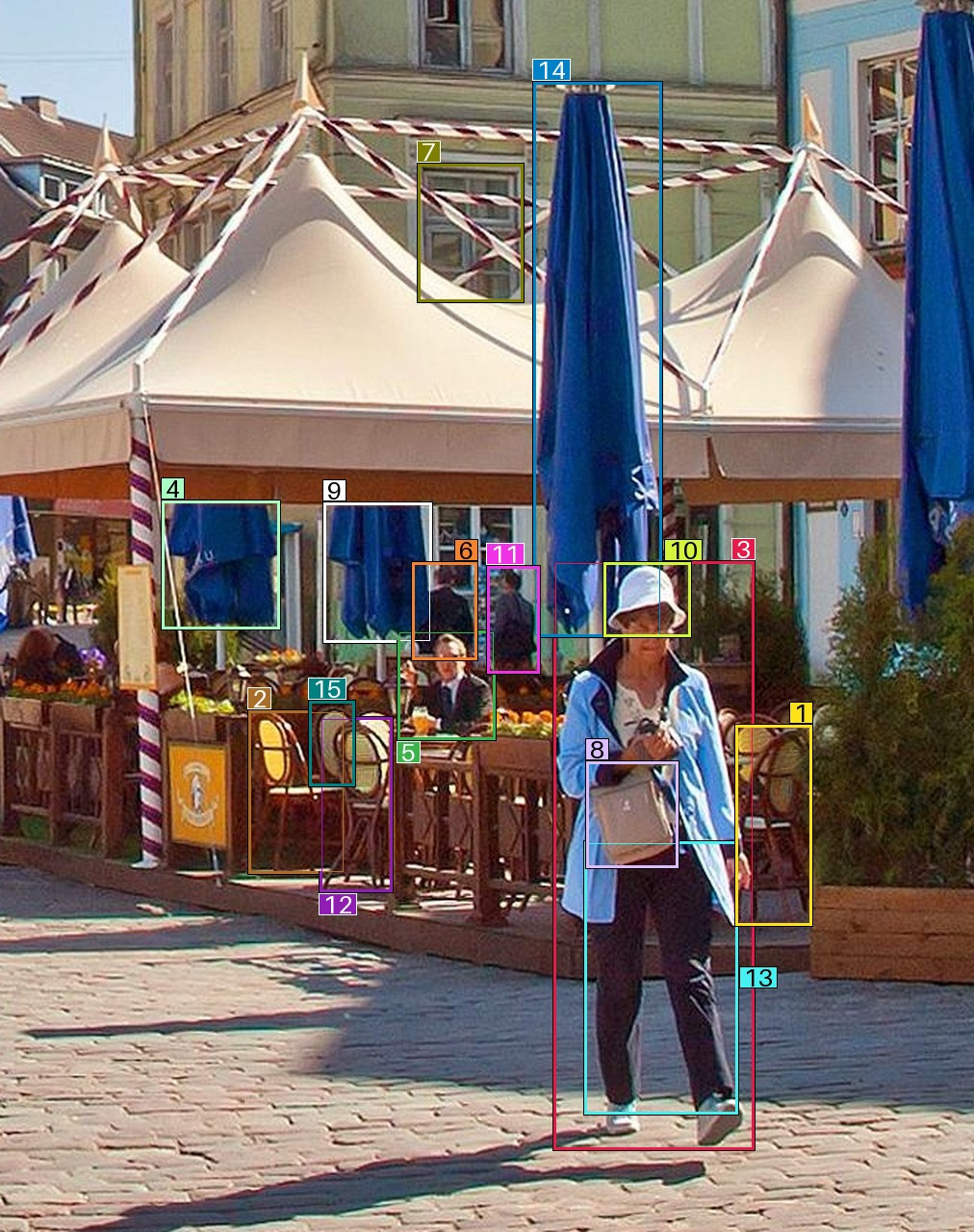}
\caption{The generator selects object 10, the hat described by the question.}
\label{fig:curation-entity-som}
\end{subfigure}
\caption{Single-object sample (continued): target identification.}
\end{figure}

\begin{figure}[p]
\centering
\begin{subfigure}{\textwidth}
\centering
\includegraphics[width=\textwidth,height=.85\textheight,keepaspectratio]{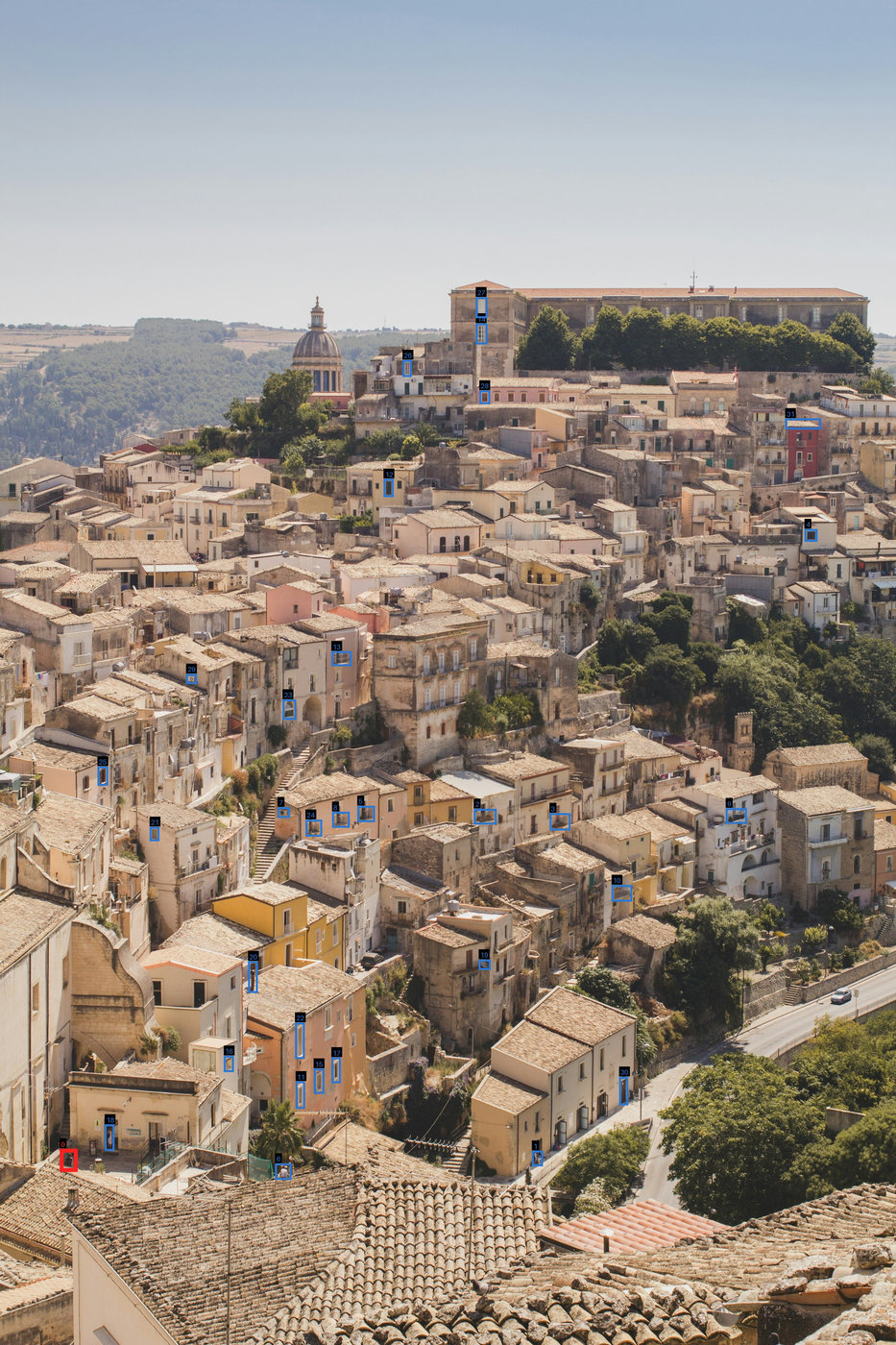}
\caption{Candidate objects detected by SAM~3. The red bounding box marks the sampled object.}
\label{fig:curation-removal-objects}
\end{subfigure}
\caption{Unanswerable sample: the sampled target object.}
\label{fig:curation-removal}
\end{figure}

\begin{figure}[p]
\ContinuedFloat
\centering
\begin{subfigure}{\textwidth}
\centering
\includegraphics[width=\textwidth]{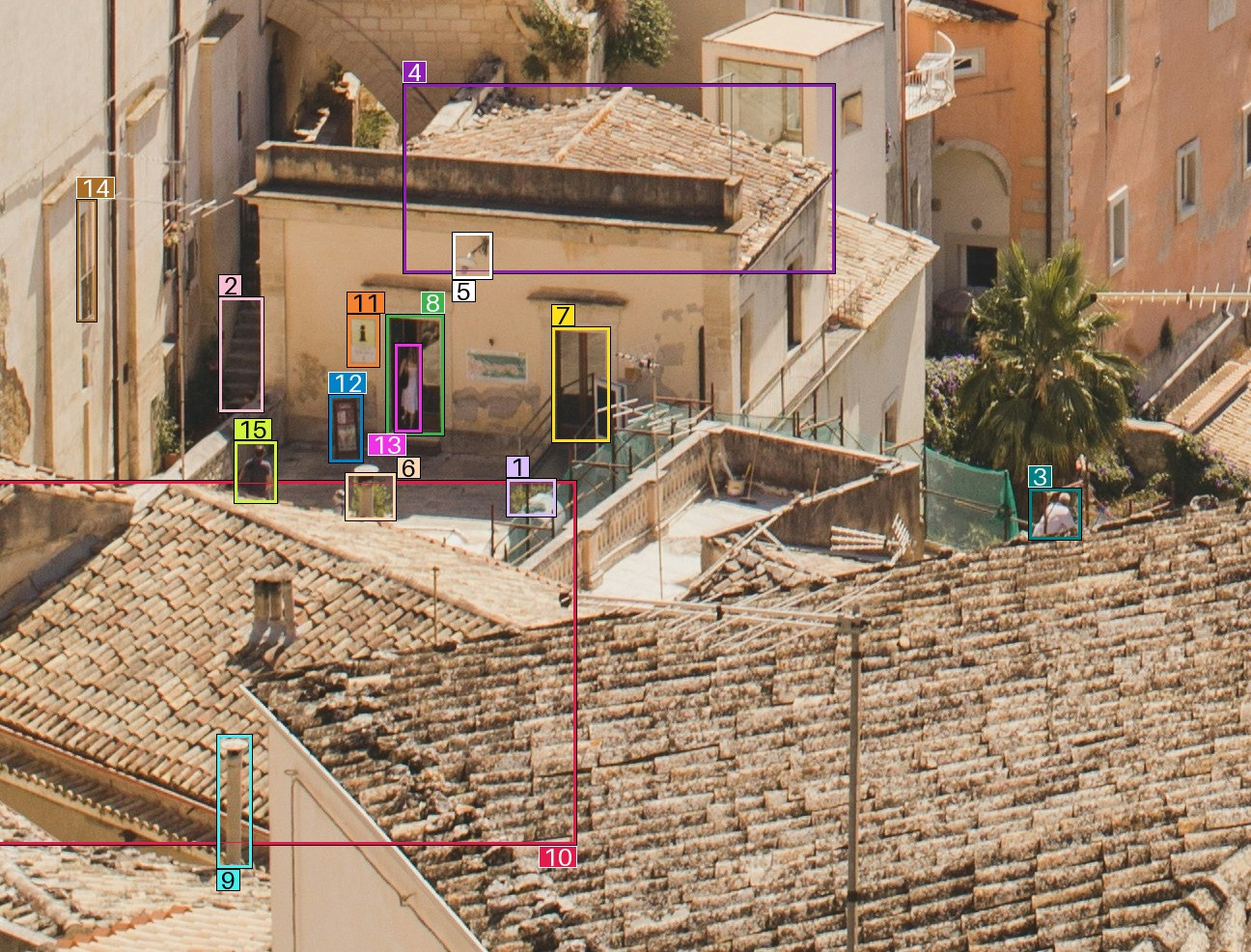}
\caption{The generator selects object 15, the person described by the question.}
\label{fig:curation-removal-som}
\end{subfigure}
\caption{Unanswerable sample (continued): identifying the target before editing.}
\end{figure}

\begin{figure}[p]
\ContinuedFloat
\centering
\begin{subfigure}{\textwidth}
\centering
\includegraphics[width=.85\textwidth]{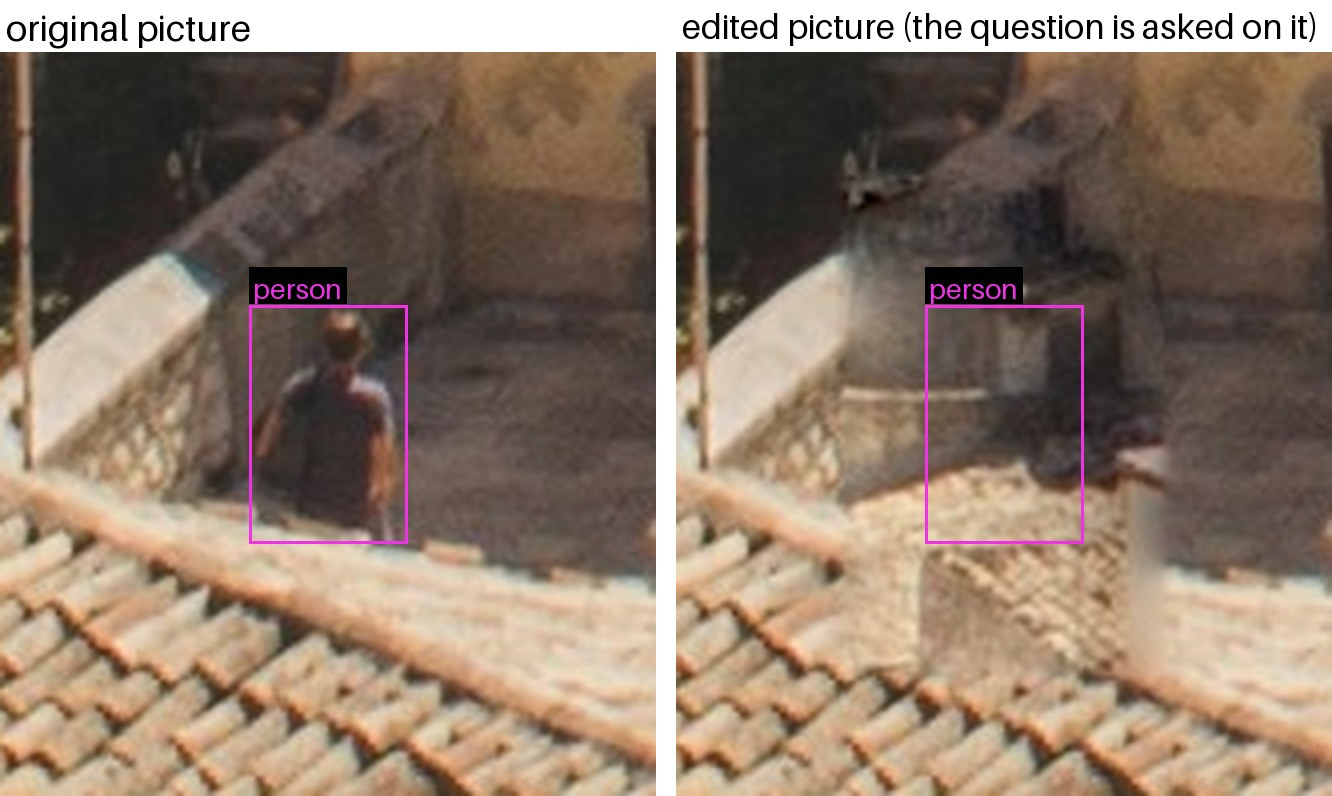}
\caption{Original evidence (left) and edited evidence (right).}
\label{fig:curation-removal-evidence}
\end{subfigure}
\caption{Unanswerable sample (continued). The edit prevents the shirt-colour comparison, but also removes the person and leaves visible seams.}
\end{figure}

\clearpage

\end{document}